\documentclass[a4paper,journal,11pt,draftclsnofoot,onecolumn]{IEEEtran}

\usepackage{dsfont}
\usepackage{amsmath}
\usepackage{amssymb}
\usepackage{amsfonts}
\usepackage{graphicx}
\usepackage{amsmath}
\usepackage[all,poly]{xy}
\usepackage{multirow}
\usepackage{algorithmic}
\usepackage{color}
\usepackage[noadjust]{cite}
\usepackage{subfigure}
\usepackage{mathrsfs}

\def\bfa{{\mathbf{a}}}

\def\bfe{{\mathbf{e}}}

\def\bfm{{\mathbf{m}}}

\def\bfu{{\mathbf{u}}}

\def\bfx{{\mathbf{x}}}

\def\bfz{{\mathbf{z}}}

\def\bfE{{\mathbf{E}}}
\def\bfF{{\mathbf{F}}}

\def\bfS{{\mathbf{S}}}

\def\bfX{{\mathbf{X}}}

\def\bfZ{{\mathbf{Z}}}

\def\calU{{\mathcal{U}}}

\def\calF{{\mathcal{F}}}
\def\calG{{\mathcal{G}}}

\def\calI{{\mathcal{I}}}

\def\calN{{\mathcal{N}}}

\def\calU{{\mathcal{U}}}

\def\bss{{\boldsymbol{s}}}

\def\bsu{{\boldsymbol{u}}}

\def\bsx{{\boldsymbol{x}}}

\def\bsz{{\boldsymbol{z}}}

\def\wtm{\widetilde{m}}
\def\wtM{\widetilde{M}}

\newcommand{\apriori}{\emph{a priori}}

\newcommand{\allobs}{\boldsymbol{\mathtt{z}}}
\newcommand{\MATobs}[1]{\bfZ_{#1}}
\newcommand{\Vobs}[1]{\bfz_{#1}}

\newcommand{\MATima}{\bfX}
\newcommand{\Vima}{\bfx}

\newcommand{\ftrans}[2]{\calF_{#1}\left(#2\right)}
\newcommand{\MATtrans}[1]{\bfF_{#1}}
\newcommand{\kernel}[1]{\boldsymbol{\kappa}_{#1}}

\newcommand{\MATnoise}[1]{\bfE_{#1}}
\newcommand{\Vnoise}[1]{\bfe_{#1}}
\newcommand{\noisevar}[1]{{s^2_{#1}}}
\newcommand{\Vnoisevar}{\bss^2}

\newcommand{\noobs}{p}
\newcommand{\nbobs}{P}

\newcommand{\nbrowobs}[1]{n_{\mathrm{x},#1}}
\newcommand{\nbcolobs}[1]{n_{\mathrm{y},#1}}
\newcommand{\nbbandobs}[1]{n_{\lambda,#1}}
\newcommand{\nbpixobs}[1]{N_{#1}}

\newcommand{\nbrowima}{m_{\mathrm{x}}}
\newcommand{\nbcolima}{m_{\mathrm{y}}}
\newcommand{\nbbandima}{m_{\lambda}}
\newcommand{\nbpixima}{M}

\newcommand{\imam}[1]{\bs{\mu}_{#1}}
\newcommand{\meansub}{\imam{\bsu}}
\newcommand{\Covsub}{\imacovmat{\bsu}}
\newcommand{\imamall}{\bar{\bs{\mu}}_{\bfu}}
\newcommand{\imacovall}{\bar{\bs{\bs{\Sigma}}}_{\bfu}}

\newcommand{\imacovmat}[1]{\bs{\Sigma}_{#1}}

\newcommand{\hypervect}{\boldsymbol{\Phi}}
\newcommand{\paramvect}{\boldsymbol{\theta}}
\newcommand{\sample}[2]{\tilde{#1}^{\left(#2\right)}}

\newcommand{\MAP}[1]{\hat{#1}_{\mathrm{MAP}}}
\newcommand{\MMSE}[1]{\hat{#1}_{\mathrm{MMSE}}}
\newcommand{\argmax}{\mathrm{arg}\max}

\newcommand{\norm}[1]{\left\|#1\right\|}

\newcommand{\R}{\mathds{R}}
\newcommand{\transp}{^T}

\newcommand{\Vzeros}[1]{\boldsymbol{0}_{#1}}
\newcommand{\Id}[1]{\textbf{I}_{#1}}
\newcommand{\Indicfun}[2]{\textbf{1}_{#1}\left(#2\right)}

\newenvironment{algogo}[1]{
\smallskip
\noindent \hrule\vspace{0.2\baselineskip} \hrule
\smallskip
\begin{small}
\refstepcounter{algo} \center{\bf \textsc{Algorithm \thealgo:}}
\\{\center{\bf #1}}
\smallskip
\flushleft
 } {
\end{small}
\bigskip
\hrule\vspace{0.2\baselineskip} \hrule
\smallskip }

\newcounter{algo}
\renewcommand{\thealgo}{\arabic{algo}}

\newcommand{\bs}{\boldsymbol}

\title{Bayesian Fusion of Multi-Band Images}

\author{\vspace{1cm}Qi Wei, Nicolas Dobigeon, Jean-Yves
Tourneret\thanks{Part of this work has been supported by the Chinese
Scholarship Council (CSC, 201206020007) and by the Hypanema ANR
Project
n$^\circ$ANR-12-BS03-003.}\\
\vspace{1cm}
\normalsize  University of Toulouse, IRIT/INP-ENSEEIHT/T\'eSA, Toulouse, France\\
\small\texttt{\{Qi.Wei,Nicolas.Dobigeon,Jean-Yves.Tourneret\}@enseeiht.fr}}

\begin{document}
\maketitle
\hyphenation{hie-rar-chi-cal as-tro-no-mi-cal}

\begin{abstract}
In this paper, a Bayesian fusion technique for remotely sensed
multi-band images is presented. The observed images are related to
the high spectral and high spatial resolution image to be recovered
through physical degradations, e.g., spatial and spectral blurring
and/or subsampling defined by the sensor characteristics. The fusion
problem is formulated within a Bayesian estimation framework. An
appropriate prior distribution exploiting geometrical
consideration is introduced. To compute the Bayesian estimator
of the scene of interest from its posterior distribution,
a Markov chain Monte Carlo algorithm is designed to generate samples
asymptotically distributed according to the target distribution. To
efficiently sample from this high-dimension distribution, a
Hamiltonian Monte Carlo step is introduced in the Gibbs
sampling strategy. The efficiency of the proposed fusion method
is evaluated with respect to several state-of-the-art fusion techniques.
In particular, low spatial resolution
hyperspectral and multispectral images are fused to produce a high
spatial resolution hyperspectral image.\end{abstract}


\newpage
\section{Introduction}
\label{sec:intro}
%Jensen, J.R.: Introductory Digital Image Processing: A Remote
%Sensing Perspective, 3rd edn. Prentice-Hall, Englewood Cliffs (2004)
The problem of fusing a high spatial and low spectral resolution
image %(i.e., multispectral image or panchromatic image)
with an auxiliary image of higher spectral but lower spatial
resolution, also known as multi-resolution image fusion, has been
explored for many years
\cite{Amro2011survey,Dou2007general,Wald1999}. When considering
remotely sensed images, an archetypal fusion task is the
pansharpening, which generally consists of fusing a high spatial
resolution panchromatic (PAN) image and low spatial resolution
multispectral (MS) image. Pansharpening has been addressed in the
image processing and remote sensing literatures for several decades
and still remains an active topic \cite{Tu2004,Aanaes2008,Joshi2010,Amro2011survey,Liu2012}.
More recently, hyperspectral (HS) imaging, which consists of acquiring a
same scene in several hundreds of contiguous spectral bands, has
opened a new range of relevant applications, such as target
detection \cite{Manolakis2002}, classification \cite{Chang2003} and
spectral unmixing \cite{Bioucas2012}. Naturally, to take advantage
of the newest benefits offered by HS images, the
problem of fusing HS and PAN images has been explored \cite{Cetin2009Merging,Licciardi2012}.
Capitalizing on decades of experience in MS pansharpening, most of
the HS pansharpening approaches merely adapt existing algorithms for
PAN and MS fusion \cite{Moeller2009,Chisense2012}. Other
methods are specifically designed to the HS pansharpening problem
(see, e.g., \cite{Winter2002resolution,Chen2012super,Licciardi2012}).
Conversely, the fusion of MS and HS images has been considered in
fewer research works and is still a challenging problem because of
the high dimension of the data to be processed.
Indeed, the fusion of MS and HS differs from traditional MS or HS pansharpening by the
fact that more spatial and spectral information is contained in
multi-band images. This additional information can be exploited to obtain a high spatial and spectral
resolution image. In practice, the spectral bands of panchromatic images
always cover the visible and infra-red spectra. However, in several practical applications,
the spectrum of MS data includes additional high-frequency spectral bands.
For instance the MS data of WorldView-3 have spectral bands in the intervals $[400\sim1750]$nm
and $[2145\sim 2365]$nm whereas the PAN data are in the range $[450\sim 800]$nm \cite{website:WV3}.
Another interesting example is the HS+MS suite (called Hyperspectral imager suite (HISUI)) that
has been developed by the Japanese Ministry of Economy, Trade, and Industry (METI) \cite{Ohgi2010}.
HISUI is the Japanese next-generation Earth-observing sensor composed of
HS and MS imagers and will be launched by H-IIA rocket in 2015
or later as one of mission instruments onboard JAXA's ALOS-3 satellite.
Some research activities have already been conducted for this practical
multi-band fusion problem \cite{Yokoya2013}. Noticeably, a lot of pansharpening methods, such
as component substitution \cite{Shettigara1992}\cite{Dou2007general}, relative spectral
contribution \cite{Zhou1998wavelet} and high-frequency injection
\cite{Gonzalez2004fusion} are inapplicable or inefficient for the
HS+MS fusion problem. To address the challenge raised by the high
dimensionality of the data to be fused, innovative methods need to
be developed. This is the main objective of this paper.

As demonstrated in \cite{Hardie2004,Zhang2009}, the fusion of HS and MS
images can be conveniently formulated within a Bayesian inference
framework. Bayesian fusion allows an intuitive interpretation of
the fusion process via the posterior distribution. Since the
fusion problem is usually ill-posed, the Bayesian
methodology offers a convenient way to regularize the problem by
defining appropriate prior distribution for the scene of interest.
Following this strategy, Hardie \emph{et al.} proposed a Bayesian
estimator for fusing co-registered high spatial-resolution MS
and high spectral-resolution HS images \cite{Hardie2004}. To improve
the denoising performance, Zhang \emph{et. al} implemented the
estimator of \cite{Hardie2004} in the wavelet domain
\cite{Zhang2009}. In \cite{Zhang2012}, Zhang \emph{et al.} derived
an expectation-maximization (EM) algorithm to maximize the posterior
distribution of the unknown image via deblurring and denoising steps.

In this paper, a prior knowledge accounting for artificial
constraints related to the fusion problem is incorporated within the model
via the prior distribution assigned to the scene to be estimated. Many strategies
related to HS resolution enhancement have been proposed to define this prior distribution.
For instance, in \cite{Joshi2010}, the highly
resolved image to be estimated is \apriori modeled by an
in-homogeneous Gaussian Markov random field (IGMRF). The parameters
of this IGMRF are empirically estimated from a panchromatic image in
the first step of the analysis. In \cite{Hardie2004} and related works
\cite{Eismann2004,Eismann2005}, a multivariate Gaussian
distribution is proposed as prior distribution for the unobserved
scene. The resulting conditional mean and covariance matrix can then
be inferred using a standard clustering technique \cite{Hardie2004}
or using a stochastic mixing model \cite{Eismann2004,Eismann2005}, incorporating spectral
mixing constraints to improve spectral accuracy in the estimated high
resolution image. In this paper, we propose to explicitly
exploit the acquisition process of the different images. More precisely, the
sensor specifications (i.e., spectral or spatial responses) are exploited
to properly design the spatial or spectral degradations suffered by the
image to be recovered \cite{Otazu2005}. Moreover, to define the prior distribution
assigned to this image, we resort to geometrical considerations well admitted in the HS
imaging literature devoted to the linear unmixing problem
\cite{Bioucas2012}. In particular, the high spatial resolution HS
image to be estimated is assumed to live in a lower dimensional
subspace, which is a suitable hypothesis when the observed scene is
composed of a finite number of macroscopic materials.

%In this work, rather than looking for a complex prior model by
%injecting possibly available \apriori information, we propose to use
%conjugate multivariate Gaussian distribution as prior distribution
%for target image. In \cite{Hardie2004}, the observation model of MS image, which is
%available and also known as spectral response , in many scenarios,
%is not explored in the formulation.

%\remark{To be clarified: In \cite{Zhang2009}, Zhang implements this
%MAP estimator in wavelet domain. Nevertheless, principal components
%analysis (PCA), which is suggested in \cite{Hardie2004}, is not
%adopted in \cite{Zhang2009}. The performance improvement of
%\cite{Zhang2009} comparing with \cite{Hardie2004} is potentially
%from the estimation of covariance matrix as for large number of
%bands, the covariance matrix may be ill-conditioned or even singular
%without decreasing the band dimensions.}

Within a Bayesian estimation framework, two statistical estimators
are generally considered. The minimum mean square error (MMSE)
estimator is defined as the mean of the posterior distribution. Its
computation generally requires intractable multidimensional
integrations. Conversely, the maximum a posteriori (MAP) estimator
is defined as the mode of the posterior distribution and is usually
associated with a penalized maximum likelihood approach. Mainly due to
the complexity of the integration required by the computation of
the MMSE estimator (especially in high-dimension data space), most of
the Bayesian estimators have proposed to solve the HS and MS fusion
problem using a MAP formulation
\cite{Hardie2004,Zhang2009,Joshi2006}. However, optimization
algorithms designed to maximize the posterior distribution may
suffer from the presence of local extrema, that prevents any
guarantee to converge towards the actual maximum of the posterior.
%While most researchers follow maximum a posterior (MAP) law
%\cite{Hardie2004}\cite{Zhang2009}\cite{Joshi2006} for optimization,
%in terms of the assessment, mean square error (MSE) is always used
%to evaluate the fusion quality. As a matter of fact, the
%\emph{minimum mean square error} (MMSE) estimation is not easy to
%evaluate as the integral of posterior density is dramatically
%cumbersome, especially in high-dimension data space.
In this paper, we propose to compute the MMSE estimator of the
unknown scene by using samples generated by a Markov chain Monte
Carlo (MCMC) algorithm. The posterior distribution resulting from
the proposed forward model and the \apriori modeling is defined in a
high dimensional space, which makes difficult the use of any
conventional MCMC algorithm, e.g., the Gibbs sampler
\cite{Casella1992} or the Metropolis-Hastings sampler
\cite{Hastings1970}. To overcome this difficulty, a particular MCMC
scheme, called Hamiltonian Monte Carlo (HMC) algorithm, is derived
\cite{Duane1987,Neal1993,Neal2010}. It differs from the standard
Metropolis-Hastings algorithm by exploiting Hamiltonian evolution
dynamics to propose states with higher acceptance ratio,
reducing the correlation between successive samples.% The MMSE and MAP estimations for unknown
%parameters can be both computed from generated samples.

The paper is organized as follows. Section \ref{sec:problem} formulates
the fusion problem in a Bayesian framework, with a
particular attention to the forward model that exploits
physical considerations. Section \ref{sec:model} derives the
hierarchical Bayesian model to obtain the joint posterior
distribution of the unknown image, its parameters and hyperparameters. In Section
\ref{sec:Gibbs}, the hybrid Gibbs sampler based on Hamiltonian MCMC
is introduced to sample the desired posterior distribution.
Simulations are conducted in Section \ref{sec:simulation} and
conclusions are finally reported in Section \ref{sec:conclusions}.

\section{Problem formulation}
\label{sec:problem}
\subsection{Notations and observation model}
Let $\MATobs{1},\ldots,\MATobs{\nbobs}$ denote a set of $\nbobs$
images acquired  by different optical sensors for a same scene
$\MATima$. These images are assumed to come from possibly
heterogeneous imaging sensors. Therefore, these measurements can be
of different natures, e.g., PAN, MS and
HS, with different spatial and/or spectral resolutions.
As in many practical situations, the observed data
$\MATobs{\noobs}$, $\noobs=1,\ldots,\nbobs$, are supposed to be
degraded versions of the high-spectral and high-spatial resolution
scene $\MATima$, according to the following observation model
\begin{equation}
\label{eq:observation}
 \MATobs{\noobs} = \ftrans{\noobs}{\MATima} + \MATnoise{\noobs}.
\end{equation}
In \eqref{eq:observation}, $\ftrans{\noobs}{\cdot}$ is a linear
transformation that models the degradation operated on $\MATima$.
As previously assumed in numerous works (see for instance
\cite{Joshi2006,Fasbender2008,Elbakary2008,Zhang2009,Joshi2010}
among some recent contributions), these degradations may include
spatial blurring, spectral blurring, decimation operation, etc. In
what follows, the remotely sensed images $\MATobs{\noobs}$
($\noobs=1,\ldots,\nbobs$) and the unobserved scene $\MATima$ are
assumed to be pixelated images of sizes $\nbrowobs{\noobs}\times
\nbcolobs{\noobs}\times \nbbandobs{\noobs}$ and $\nbrowima\times
\nbcolima\times \nbbandima$, respectively, where
$\cdot_{\mathrm{x}}$ and $\cdot_{\mathrm{y}}$ refer to both spatial
dimensions of the images, and $\cdot_{\lambda}$ is for the spectral
dimension. Moreover, in the right-hand side of \eqref{eq:observation}, $\MATnoise{\noobs}$
stands for an additive error term that both reflects the mismodeling
and the observation noise.

%Note that, for
%simplicity reasons but without any loss of generality, all the
%observed images $\MATobs{\noobs}$ share the same spatial dimensions
%as the high resolution scene $\MATima$.

Classically, the observed image $\MATobs{\noobs}$ can be
lexicographically ordered to build the $\nbpixobs{\noobs} \times 1$
vector $\Vobs{\noobs}$, where $\nbpixobs{\noobs}=\nbrowobs{\noobs}
\nbcolobs{\noobs} \nbbandobs{\noobs}$ is the total number of
measurements in the observed image $\MATobs{\noobs}$. For multi-band
images, this vectorization can be performed following either band
sequential (BSQ), band interleaved by line (BIL) or band interleaved
by pixel (BIP) schemes (see \cite[pp. 103--104]{Campbell2002} for a
more detailed description of these data format conventions). For
writing convenience, but without any loss of generality, the BIP-like
vectorization scheme is adopted in what follows (see paragraph
\ref{subsubsec:scene_prior}). As a consequence, the observation
equation \eqref{eq:observation} can be easily rewritten as follows
\begin{equation}
\label{eq:observation2}
  \Vobs{\noobs} = \MATtrans{\noobs}{\Vima} + \Vnoise{\noobs}
\end{equation}
where the $\nbpixima \times 1$ vector $\Vima$ and the
$\nbpixobs{\noobs} \times 1$ vector $\Vnoise{\noobs}$ are
ordered versions of the scene $\MATima$ (with $\nbpixima= \nbrowima
\nbcolima \nbbandima$) and the noise term $\MATnoise{\noobs}$,
respectively. In this work, the noise vector $\Vnoise{\noobs}$ will be assumed to
be a band-dependent Gaussian sequence,
%i.e., $\Vnoise{\noobs} \sim
%\calN\left(\Vzeros{\nbpixobs{\noobs}},\noisevar{\noobs}\Id{\nbpixobs{\noobs}}\right)$
i.e., $\Vnoise{\noobs} \sim \calN\left(\Vzeros{\nbpixobs{\noobs}},{\bf\Lambda}_p\right)$
%$\noisevar{\noobs}\Id{\nbpixobs{\noobs}}$
where $\Vzeros{\nbpixobs{\noobs}}$ is an $\nbpixobs{\noobs}\times
1$ vector made of zeros and ${\bf\Lambda}_p = \Id{\nbrowobs{\noobs}\nbcolobs{\noobs}} \otimes \bfS_p$ is an
$\nbpixobs{\noobs}\times \nbpixobs{\noobs}$ matrix where $ \Id{\nbrowobs{\noobs}\nbcolobs{\noobs}} \in \mathbb{R}^{
\nbrowobs{\noobs}\nbcolobs{\noobs} \times \nbrowobs{\noobs}\nbcolobs{\noobs}}$ is
the identity matrix, $\otimes$ is the Kronecker product and $\bfS_p \in \mathbb{R}^{\nbbandobs{\noobs} \times \nbbandobs{\noobs}}$
is a diagonal matrix containing the noise variances, i.e.,
$\bfS_p = \textrm{diag} \left[\noisevar{\noobs,1} \cdots \noisevar{\noobs,\nbbandobs{\noobs}}\right]$.
The Gaussian noise assumption is quite popular in
image processing \cite{Jalobeanu2004,Duijster2009,Xu2011} as it
facilitates the formulation of the likelihood and the optimization
algorithm. However, the proposed Bayesian model could be modified,
for instance to take into account correlations between
spectral bands, following the strategy in \cite{Dobigeon2008icassp}.
Note also that the variance matrix $\bfS_p$ of the noise vector
$\Vnoise{\noobs}$ depends on the observed data $ \Vobs{\noobs}$,
since the signal-to-noise ratio may differ from one
sensor to another.

In \eqref{eq:observation2}, $\MATtrans{\noobs}$ is an
$\nbpixobs{\noobs} \times \nbpixima$ matrix that reflects the
spatial and/or spectral degradation $\ftrans{\noobs}{\cdot}$
operated on $\Vima$. As in \cite{Hardie2004},
$\ftrans{\noobs}{\cdot}$ can represent a spatial decimating
operation. For instance, when applied to a single-band image (i.e.,
$\nbbandobs{\noobs}=\nbbandima=1$) with a decimation factor $q$ in
both spatial dimensions, it is easy to show that $\MATtrans{\noobs}$
is an $\nbrowobs{\noobs} \nbcolobs{\noobs} \times
\nbrowima\nbcolima$ block diagonal matrix given in \eqref{eq:Fp}
 with $\nbrowima = d \nbrowobs{\noobs}$ and
$\nbcolima = d \nbcolobs{\noobs}$ \cite{Schultz1994}.

\begin{equation}
\label{eq:Fp}
  \MATtrans{\noobs} = \frac{1}{d^2}\left(
                                     \begin{array}{ccccc}
                                       \underbrace{1  1  \ldots     1}_{d^2} &   &                    &           &                    \\
                                                         &   & \underbrace{1  1  \ldots     1}_{d^2} &           &                    \\
                                                         &   &                    & \ddots    &                    \\
                                                         &   &                    &           & \underbrace{1  1  \ldots     1}_{d^2} \\
                                     \end{array}
                                   \right)
\end{equation}

Another example of degradation frequently encountered in the signal and
image processing literature is spatial blurring \cite{Zhang2009},
where $\ftrans{\noobs}{\cdot}$ usually represents a
$2$-dimensional convolution by a kernel $\kernel{\noobs}$.
Similarly, when applied to a single-band image, $\MATtrans{\noobs}$
is an $n_x n_y \times n_x n_y$ (generally sparse) Toeplitz matrix,
that is symmetric for a symmetric convolution kernel
$\kernel{\noobs}$.

The problem addressed in this paper consists of recovering the
high-spectral and high-spatial resolution scene $\Vima$ by fusing
the various spatial and/or spectral information provided by all the
observed images $\allobs =\left\{\Vobs{1},\ldots,\Vobs{\nbobs}\right\}$.
To facilitate reading, notations have been summarized in Table \ref{tb:term}.

\begin{table*}
\centering \caption{Notations}
\begin{tabular}{c|c|c}
\hline
\textbf{Notation} & \textbf{Definition} & \textbf{Size} \\
\hline
$\MATima$  & unobserved scene  or target image & $\nbrowima\times\nbcolima\times \nbbandima$\\
$\Vima$    & vectorization of $\MATima$        & $\nbrowima\nbcolima\nbbandima \times  1$\\
$\bsx_{i}$ & band vector at $i$th position of $\Vima$                 &$\nbbandima \times  1$\\
$\bfu$     & vectorized image after reducing band dimension by PCA    &$\nbrowima\nbcolima\wtm_{\lambda} \times  1$\\
$\bsu_{i}$ & band vector at $i$th position of $\bfu$                  &$\wtm_{\lambda} \times  1$\\
$\imamall$ & prior mean of $\bfu$       & $\nbrowima\nbcolima\wtm_{\lambda} \times  1$\\
$\imacovall$  & prior covariance of $\bfu$ & $\nbrowima\nbcolima\wtm_{\lambda} \times  \nbrowima\nbcolima\wtm_{\lambda}$\\
$\imam{\bsu_i}$ & prior mean of $\bsu_{i}$       & $\wtm_{\lambda} \times  1$\\
$\imacovmat{\bsu_i}$  & prior covariance of $\bsu_{i}$ & $\wtm_{\lambda} \times  \wtm_{\lambda}$\\
$\MATobs{p}$ &$p$th remotely sensed images  & $\nbrowobs{\noobs} \times \nbcolobs{\noobs} \times \nbbandobs{\noobs} $\\
$\Vobs{p}$ & vectorization of $\MATobs{p}$  & $\nbrowobs{\noobs}\nbcolobs{\noobs}\nbbandobs{\noobs} \times  1$\\
$\allobs$  & set of $P$ vectorized observed images $\Vobs{p}$ &
$\nbrowobs{\noobs}\nbcolobs{\noobs}\nbbandobs{\noobs}P \times 1  $\\
 \hline
\end{tabular}
\label{tb:term}
\end{table*}

\subsection{Bayesian estimation of $\Vima$}
In this work, we propose to estimate the unknown scene $\Vima$
within a Bayesian estimation framework. In this statistical estimation
scheme, the fused highly-resolved image
$\Vima$ is inferred through its posterior distribution
$f\left(\Vima|\allobs\right)$. Given the observed data, this
target distribution can be derived from the likelihood function
$f\left(\allobs|\Vima\right)$ and the prior distribution
$f\left(\Vima\right)$ by using the Bayes' formula
\begin{equation}
\label{eq:Bayes_rule}
  f\left(\Vima|\allobs\right) =
  \frac{f\left(\allobs|\Vima\right)f\left(\Vima\right)}{f\left(\allobs\right)}.
\end{equation}
%Generally, the normalizing term $f\left(\allobs\right)$, sometimes
%called evidence, is not easily calculable as it is obtained after
%marginalizing the joint distribution $
%f\left(\allobs,\Vima\right)=f\left(\allobs|\Vima\right)f\left(\Vima\right)$
%over the unknown quantity $\Vima$
%\begin{equation}
%  f\left(\allobs\right) = \int
%  f\left(\allobs|\Vima\right)f\left(\Vima\right)
%  d\Vima.
%\end{equation}
Based on the posterior distribution \eqref{eq:Bayes_rule}, several
estimators of the scene $\Vima$ can be investigated. For instance,
maximizing $f\left(\Vima|\allobs\right)$ leads to the MAP estimator $\MAP\Vima$
\begin{equation}
\begin{split}
\label{eq:MAP}
  \MAP\Vima &= \argmax_{\Vima} f\left(\Vima|\allobs\right)\\
              &=\argmax_{\Vima}
              f\left(\allobs|\Vima\right)f\left(\Vima\right).
\end{split}
\end{equation}
This estimator has been widely exploited for HS image
resolution enhancement (see for instance
\cite{Hardie2004,Eismann2004,Eismann2005} or more recently
\cite{Joshi2010,Zhang2009}). This work proposes to focus on the
first moment of the posterior distribution
$f\left(\Vima|\allobs\right)$, which is known as the posterior mean
estimator or the \emph{minimum mean square error} estimator
$\MMSE\Vima$. This estimator is defined as
\begin{equation}
\begin{split}
\label{eq:MMSE}
  \MMSE\Vima &= \mathrm{E}\left[\Vima|\allobs\right] \\
               & = \int \Vima
               f\left(\Vima|\allobs\right)d\Vima\\
               & = \frac{\int \Vima
               f\left(\allobs|\Vima\right)f\left(\Vima\right)d\Vima}{\int
               f\left(\allobs|\Vima\right)f\left(\Vima\right)d\Vima}.
\end{split}
\end{equation}
In this work, we propose a flexible and relevant statistical model
to solve the fusion problem. Deriving the corresponding Bayesian estimators %$\MAP\Vima$ and
$\MMSE\Vima $ defined in \eqref{eq:MAP} and \eqref{eq:MMSE},
requires the definition of
the likelihood function $ f\left(\allobs|\Vima\right)$ and the
prior distribution  $f\left(\Vima\right)$. These quantities
are detailed in the next section.

\section{Hierarchical Bayesian model}
\label{sec:model}
\subsection{Likelihood function}
The statistical properties of the noise vectors $\Vnoise{\noobs}$
($\noobs=1,\ldots,\nbobs$) allow one to state that the observed
vector $\Vobs{\noobs}$ is normally distributed with mean vector
$\MATtrans{\noobs}\Vima$ and covariance matrix
${\bf\Lambda}_p$. Consequently, the
likelihood function, that represents a data fitting term
relative to the observed vector $\Vobs{\noobs}$, can be easily
derived leading to
\begin{equation}
\begin{split}
\label{eq:likelihood_marginal}
  f\left(\Vobs{\noobs}|\Vima,{\bf\Lambda}_p\right) &=
  \left(2\pi\right)^{-\frac{\nbpixobs{\noobs}}{2}} \vert{\bf\Lambda}_p\vert^{-\frac{\nbrowobs{\noobs}\nbcolobs{\noobs}}{2}} \\
  &\times \exp\left(-\frac{1}{2}\left(\Vobs{\noobs}-\MATtrans{\noobs}\Vima\right)^T {\bf\Lambda}_p^{-1} \left(\Vobs{\noobs}-\MATtrans{\noobs}\Vima\right)\right)
\end{split}
\end{equation}
where $\vert{\bf\Lambda}_p\vert$ is the determinant of the matrix ${\bf\Lambda}_p$.
%$\norm{\Vima}=\left(\Vima\transp\Vima\right)^{\frac{1}{2}}$ is the $\ell_2$-norm.
As mentioned in the previous section, the collected measurements $\allobs$ may have been acquired by different
(possibly heterogeneous) sensors. Therefore, the observed vectors
$\Vobs{1},\ldots,\Vobs{\nbobs}$ can be generally assumed to be
independent, conditionally upon the unobserved scene $\Vima$ and the
noise  covariances ${\bf\Lambda}_1,\ldots,{\bf\Lambda}_p$. As a
consequence, the joint likelihood function of the observed data is
\begin{equation}
\begin{split}
  \label{eq:likelihood_joint}
    f&\left(\allobs|\Vima,\bf\Lambda\right) = \prod_{\noobs=1}^{\nbobs}
    f\left(\Vobs{\noobs}|\Vima,{\bf\Lambda}_p\right)\\
\end{split}
\end{equation}
with
%& = \left[\prod_{\noobs=1}^{\nbobs}\left(\frac{1}{2\pi\noisevar{\noobs}}\right)^{\frac{\nbpixobs{\noobs}}{2}}\right]
%         \exp\left(-\sum_{\noobs=1}^{\nbobs}\frac{\norm{\Vobs{\noobs}-\MATtrans{\noobs}\Vima}^2}{2\noisevar{\noobs}}\right)
${\bf\Lambda}=\left({\bf\Lambda}_1,\ldots,{\bf\Lambda}_P \right)^T$.

\subsection{Prior distributions}
\label{sec:prior} The unknown parameters are the scene $\Vima$ to be
recovered and the noise covariance matrix $\bf\Lambda$ relative to
each observation. In this section, prior distributions are
introduced for these parameters. %$\paramvect=\left\{\Vima,\Vnoisevar\right\}$

\subsubsection{Scene prior}
\label{subsubsec:scene_prior}

Following a BIP strategy, the vectorized image $\Vima$ can be
decomposed as $\Vima =
\left[\bsx_1^T,\bsx_2^T,\cdots,\bsx_{\nbrowima\nbcolima}^T\right]^T$,
where $\bsx_i=\left[x_{i,1},x_{i,2},\cdots,x_{i,\nbbandima}\right]^T
$ is  the $\nbbandima \times 1$ vector corresponding to the $i$th
spatial location (with $i=1,\cdots,\nbrowima\nbcolima$). The
HS vector $\bsx_i$ usually lives in a subspace whose
dimension is much smaller than the number of bands $\nbbandima$
\cite{Chang1998,Bioucas2008}. In order to account for this subspace
of reduced dimension $\wtm_{\lambda}$, we introduce a linear
transformation from $\mathbb{R}^{\nbbandima \times 1}$ to
$\mathbb{R}^{\wtm_{\lambda} \times 1}$ such that
\begin{equation}
\label{eq:subspace}
\bsu_i= \mathbf{V} \bsx_i
\end{equation}
where $\bsu_i$ is the projection of the vector $\bsx_i$  onto
the subspace of interest and the transformation matrix $\mathbf{V}$
is of size $\wtm_{\lambda} \times \nbbandima$. Using the notation
$\bfu = \left[\bsu_1^T,\bsu_2^T,\cdots,\bsu_{\nbrowima\nbcolima}^T\right]^T$,
we have $\mathbf{u}=\mathfrak{V}\Vima$, where $\mathfrak{V}$ is an
$\wtM \times M$ block-diagonal matrix whose blocks are equal to
$\mathbf{V}$ and  $\wtM = \nbrowima \nbcolima \wtm_{\lambda}$.
Instead of assigning a prior distribution to the vectors
$\bsx_i$, we propose to define a prior for the projected vectors
$\bsu_i$ ($i=1,\cdots,\nbrowima \nbcolima$)
%Note $\textrm{diag}[\bfA]$ represents constructing a block diagonal matrix whose blocks are equal to  $\mathbf{V}$ diagonal is $\bfA$ and $\mathfrak{V}$ is of size
%$\wtm \times M$ ,where $\wtm= \nbrowima \nbcolima \widetilde{\nbbandima}$.
%Therefore, a prior distribution is assumed to $\Vima$ in the subspace $\mathbb{R}^{\widetilde{\nbbandima} \times 1}$
%as is shown in \eqref{eq:prior_scene}.
% For the subspace learning, there exists  numerous of methods \cite{Cai2005,Cai2007,Bioucas2008}.
%\textcolor{red}{}
% $\textcolor{red}{\wtm_{\lambda}}$
%In this work, we don't assume prior spatial dependency between adjacent pixels. $\bsx_i$ is assigned to obey independent identical distribution (i.i.d)
%\textcolor{red}{}
\begin{equation}
\label{eq:prior_scene}
  \bsu_i | \imam{\bsu_i},\imacovmat{\bsu_i} \sim
  \calN\left(\imam{\bsu_i},\imacovmat{\bsu_i}\right).
\end{equation}
%for , where $\meansub =
%\left(\mu_1,\mu_2,\cdots,\mu_{\wtm_{\lambda}}\right)^T$ and $\Covsub
%= $
%\begin{equation}
%\label{eq:prior_cov}
%\Covsub=   \left(
%                       \begin{array}{cccc}
%                       \sigma^2_{1,1} & \sigma^2_{1,2} & \cdots  & \sigma^2_{1,\wtm_{\lambda}}      \\
%                       \sigma^2_{2,1} & \sigma^2_{2,2} & \cdots  & \sigma^2_{2,\wtm_{\lambda}}    \\
%                          \vdots      & \vdots         & \ddots  & \vdots                       \\
%                       \sigma^2_{\wtm_{\lambda},1} & \sigma^2_{\wtm_{\lambda},2} & \cdots & \sigma^2_{\wtm_{\lambda},\wtm_{\lambda}}   \\
%                       \end{array}
%                    \right).
%\end{equation}
Assigning a prior to the projected vectors $\bsu_i$ allows the ill-posed problem
\eqref{eq:observation2} to be regularized. The covariance matrix $\imacovmat{\bsu_i}$
is designed to explore the correlations between the different spectral bands after projection
in the subspace of interest. Also, the mean $\imamall$ of the whole
image $\bfu$ as well as its covariance matrix $\imacovall$ can be
constructed from $\imam{\bsu_i}$ and $\imacovmat{\bsu_i}$ as follows
\begin{equation}
\label{eq:para_whole}
\begin{split}
\imamall=\big[\imam{\bsu_{1}}^T,\cdots,\imam{\bsu_{\nbrowima\nbcolima}}^T\big]^T, \\
\imacovall =
\textrm{diag}\big[\imacovmat{\bsu_1},\cdots,\imacovmat{\bsu_{\nbrowima\nbcolima}}\big].
\end{split}
\end{equation}
%\imamall=\big[\underbrace{\imam{(1)T},\imam{(2)T},\cdots,\imam{(\nbrowima\nbcolima)T}}_{\nbrowima\nbcolima}\big]^T, \\
%\imacovall =
%\textrm{diag}\big[\underbrace{\imacovmat{1},\imacovmat{2},\cdots,\imacovmat{\nbrowima\nbcolima}}_{\nbrowima\nbcolima}\big].
Note that the choice of the hyperparameters $\imamall$ and $\imacovall$ will
be discussed later in Section \ref{subsec:hyper_prior}.
Choosing a Gaussian prior for the vectors $\bsu_i$ is motivated by
the fact this kind of prior has been used successfully in several
works related to the fusion of multiple degraded images, including
\cite{Hardie1997,Eismann2004,Woods2006}. Note that the Gaussian
prior has also the interest of being a conjugate distribution
relative to the statistical model in \eqref{eq:likelihood_joint}. As
it will be shown in Section \ref{sec:Gibbs}, coupling this Gaussian
prior distribution with the Gaussian likelihood function leads to
simpler estimators constructed from the posterior distribution
$f\left(\bfu|\allobs\right)$. Finally, it is interesting to mention that
the proposed method is quite robust to the non-Gaussianity of the image.
Some additional results obtained for synthetic non-Gaussian images as well as
related discussions are available in \cite{Qi2014_TechRep}.

\subsubsection{Noise variance priors}
As in numerous works including \cite{Punskaya2002},
conjugate inverse-gamma distributions are chosen as prior
distributions for the noise variances $\noisevar{\noobs,i}$
($i=1,\ldots,\nbbandobs{\noobs}, \noobs=1,\ldots,\nbobs$)
\begin{equation}
  \label{eq:prior_noisevar}
    \noisevar{\noobs,i} | \nu,\gamma\sim
\calI\calG\left(\frac{\nu}{2},\frac{\gamma}{2}\right).
\end{equation}

Again, these conjugate distributions will allow closed-form expressions to be
obtained for the conditional distributions $f\left(\noisevar{\noobs,i}|\allobs\right)$
of the noise variances. Other motivations for using this kind of prior
distribution can be found in \cite{Gelman2006prior}.
In particular, the inverse-gamma distribution is a very flexible distribution whose shape
can be adjusted by its two parameters. For simplicity, we propose to fix the hyperparameter $\nu$ whereas
the hyperparameter $\gamma$ will be estimated from the data. By assuming the
variances $\Vnoisevar= \left[\noisevar{1,1},\ldots,\noisevar{1,\nbbandobs{1}},\ldots,\noisevar{P,1},\ldots,\noisevar{P,\nbbandobs{P}}\right]$
are \apriori independent, the joint prior distribution of the noise variance
vector $\Vnoisevar$ is
\begin{equation}
  \label{eq:prior_Vnoisevar}
    f\left(\Vnoisevar | \nu,\gamma\right) = \prod_{\noobs=1}^{\nbobs} \prod_{i=1}^{\nbbandobs{\noobs}}  f\left(\noisevar{\noobs,i} | \nu,\gamma\right).
\end{equation}

\subsection{Hyperparameter priors}
\label{subsec:hyper_prior}
The hyperparameter vector associated with the parameter priors defined
above includes $\imamall$, $\imacovall$ and $\gamma$. The quality of the fusion algorithm investigated
in this paper depends on the values of the hyperparameters that need to be adjusted carefully.
Instead of fixing all these hyperparameters {\apriori}, we propose to estimate some of them from the data by using
a hierarchical Bayesian algorithm \cite[Chap. 8]{Robert2007}. Specifically, we propose to fix $\imamall$ as the interpolated
HS image in the subspace of interest following the strategy in \cite{Hardie2004}. Similarly, to reduce the number of statistical
parameters to be estimated, all the covariance matrix are assumed to be equal, i.e., $\imacovmat{\bsu_1}=\cdots=\imacovmat{\bsu_{\nbrowima\nbcolima}}=\Covsub$.
Thus, the hyperparameter vector to be estimated jointly with the parameters of interest is $\hypervect=\left\{\Covsub,\gamma\right\}$.
The prior distributions for these two hyperparameters are defined below.

%\subsubsection{Hyperparameter $\meansub$}
%The hyperparameter $\meansub$ is assigned a conjugate Gaussian distribution
%\begin{equation}
%\label{eq:Prior_mean}
%\meansub \sim \calN \left(\imam{0}, \imacovmat{0} \right)
%\end{equation}
%where  $\imam{0}$ and $\imacovmat{0}$ are fixed to ensure a non-informative prior.% for $\meansub$.

\subsubsection{Hyperparameter $\Covsub$}
Assigning an \apriori inverse-Wishart distribution to the covariance
matrix of a Gaussian vector has provided interesting results in the
signal and image processing literature \cite{Bidon2008,Bouriga2012}.
Following these works, we have chosen the following prior for
$\Covsub$
\begin{equation}
\label{eq:Wishart} \Covsub \sim \mathcal{W}^{-1}({\mathbf\Psi},\eta)
\end{equation}
whose density is
\begin{equation*}
f(\Covsub|\mathbf\Psi,\eta) =
\frac{\left|{\mathbf\Psi}\right|^{\frac{\eta}{2}}}{2^{\frac{\eta
\wtm_{\lambda}}{2}}\Gamma_{\wtm_{\lambda}}(\frac{\eta}{2})}\left|\Covsub\right|^{-\frac{\eta+\wtm_{\lambda}+1}{2}}e^{-\frac{1}{2}\operatorname{tr}({\mathbf\Psi}\Covsub^{-1})}.
\end{equation*}
Again, the hyper-hyperparameters $\mathbf\Psi$ and $\eta$ will be fixed to provide a
non-informative prior. %for $\Covsub$.
%Note that the inverse-Wishart
%distribution reduces to the inverse-gamma distribution when
%$\wtm_{\lambda} = 1$.

%\begin{equation}
%  \Covsub \sim
%\calI\calG\left(\frac{\nu}{2},\frac{\mathbf\Psi}{2}\right).
%\end{equation}

\subsubsection{Hyperparameter $\gamma$}
To reflect the absence of prior knowledge regarding the mean noise
level, a non-informative Jeffreys' prior is assigned to the
hyperparameter $\gamma$ %\cite{Jeffreys1961}
\begin{equation}
  \label{eq:prior_gamma}
  f\left(\gamma\right) \propto
  \frac{1}{\gamma} \Indicfun{\R^+}{\gamma}
\end{equation}
where $\Indicfun{\R^+}{\cdot}$ is the indicator function defined on
$\R^+$
\begin{equation}
  \Indicfun{\R^+}{u}=\left\{
                 \begin{array}{ll}
                   1, & \hbox{if $u\in\R^+$,} \\
                   0, & \hbox{otherwise.}
                 \end{array}
               \right.
\end{equation}
The use of the improper distribution \eqref{eq:prior_gamma} is classical and can be justified  by
different means (e.g., see \cite{Gelman2006prior}), providing that the corresponding full posterior distribution is
statistically well defined, which is the case for the proposed fusion model. %(see Section~\ref{sec:Gibbs}).

\subsection{Inferring the highly-resolved HS image from the posterior distribution of its projection $\bfu$}
Following the parametrization in the prior model
\eqref{eq:subspace}, the unknown parameter vector
$\paramvect=\left\{\bfu,\Vnoisevar\right\}$ is composed of the
projected scene $\bfu$ and the noise variance vector $\Vnoisevar$. The joint
posterior distribution of the unknown parameters and hyperparameters
can be computed following the hierarchical model
\begin{equation}
\label{eq:posterior_joint}
  f\left(\paramvect,\hypervect|\allobs\right) \propto f\left(\allobs|\paramvect\right)f\left(\paramvect|\hypervect\right)f\left(\hypervect\right).
\end{equation}
By assuming prior independence between the hyperparameters
 $\Covsub$ and $\gamma$ and the parameters $\bfu$ and
$\Vnoisevar$ conditionally upon ($\Covsub,\gamma$), the
following results can be obtained
\begin{equation}
 f\left(\paramvect|\hypervect\right)=f\left(\bfu|\Covsub\right)f\left(\Vnoisevar|\gamma\right)
\end{equation}
and
\begin{equation}
 f\left(\hypervect\right)=f\left(\Covsub\right)f\left(\gamma\right).
\end{equation}
Note that $f\left(\allobs|\paramvect\right)$,
$f\left(\bfu|\Covsub\right)$ and $f\left(\Vnoisevar|\gamma\right)$
have been defined in \eqref{eq:likelihood_joint},
\eqref{eq:prior_scene} and \eqref{eq:prior_Vnoisevar}, respectively.

The posterior distribution of the projected highly resolved image
$\bfu$, required to compute the Bayesian estimators \eqref{eq:MAP}
and \eqref{eq:MMSE}, is obtained by marginalizing out the
hyperparameter vector $\hypervect$ and the noise variances
$\Vnoisevar$ from the joint posterior distribution
$f\left(\paramvect,\hypervect|\allobs\right)$
\begin{equation}
\begin{split}
\label{eq:posterior_Vima}
  f&\left(\bfu|\allobs\right) \propto \int
f\left(\paramvect,\hypervect|\allobs\right) d\hypervect
d\noisevar{1,1},\ldots,d\noisevar{P,\nbbandobs{P}}.\\
\end{split}
\end{equation}
% &\propto
%\prod_{\noobs=1}^{\nbobs}\left(\norm{\Vobs{\noobs}-\MATtrans{\noobs}\Vima}^2\right)^{-\frac{\nbpixobs{\noobs}}{2}}
%    \left(\mathbf\Psi + \norm{\Vima}^2\right)^{-\frac{\nu+\nbpixima}{2}}.
The posterior distribution \eqref{eq:posterior_Vima} is too complex
to obtain closed-form expressions of the MMSE and MAP estimators
$\MMSE\bfu$ and $\MAP\bfu$. As an alternative, this paper proposes
to use an MCMC algorithm to generate a collection of $N_{\textrm{MC}}$
samples
\begin{equation}
  \calU =
\left\{\sample{\bfu}{1},\ldots,\sample{\bfu}{N_{\textrm{MC}}}\right\}
\end{equation}
that are asymptotically distributed according to the posterior of
interest $ f\left(\bfu|\allobs\right)$. These samples will be used
to compute the Bayesian estimators of $\bfu$. More precisely, the
MMSE estimator of $\bfu$ will be approximated by an empirical
average of the generated samples
\begin{equation}
  \MMSE\bfu \approx \frac{1}{N_{\textrm{MC}}-N_{\textrm{bi}}}
\sum_{t=N_{\textrm{bi}}+1}^{N_{\textrm{MC}}} \sample{\bfu}{t}
 \label{eq:MMSE_X}
\end{equation}
where $N_{\textrm{bi}}$ is the number of burn-in iterations.
%Conversely, the MAP estimator \eqref{eq:MAP} will be approximated
%by retaining among the collection $\calX$ the vector that maximizes
%the distribution \eqref{eq:posterior_Vima}, i.e,
%\begin{equation}
%  \MAP\Vima \approx \argmax_{\Vima\in\calX} f\left(\Vima|\allobs\right).
%\end{equation}
%Note that \eqref{eq:posterior_Vima} is required to derive the MAP estimator of $\Vima$.
Once the MMSE estimate $\MMSE\bfu$ has been computed, the
highly-resolved HS image can be computed as
\begin{equation}
 \MMSE\Vima =\mathfrak{V}^T \MMSE\bfu.
\end{equation}
Sampling directly according to the marginal posterior distribution
$f\left(\bfu|\allobs\right)$ is not straightforward. Instead, we
propose to sample according to the joint posterior $f\left(\bfu,
\Vnoisevar, \Covsub|\allobs\right)$ (hyperparameter $\gamma$ has been marginalized)
by using a Metropolis-within-Gibbs sampler, which can be easily implemented
since all the conditional distributions associated with
$f\left(\bfu, \Vnoisevar, \Covsub|\allobs\right)$ are
relatively simple. The resulting hybrid Gibbs sampler is detailed in
the following section.

\section{Hybrid Gibbs Sampler}
\label{sec:Gibbs}

The Gibbs sampler has received a considerable attention in the
statistical community (see \cite{Casella1992,Robert2007}) to solve
Bayesian estimation problems. The interesting property of this Monte
Carlo algorithm is that it only requires to determine the
conditional distributions associated with the distribution of
interest. These conditional distributions are generally easier to
simulate than the joint target distribution. The block Gibbs sampler
that we propose to sample according to $f\left(\bfu,
\Vnoisevar, \Covsub|\allobs\right)$ is defined by a
$3$-step procedure reported in Algo. \ref{algo:Gibbs}. The
distribution involved in this algorithm are detailed below.

%Thus, one simulates $n$ random
%variables sequentially from the $n$ uni-variate conditionals rather than generating
%a single $n$-dimensional vector in a single pass using the full joint distribution.

\begin{figure}[h!]
    \begin{algogo}{Hybrid Gibbs sampler}
    \label{algo:Gibbs}
    \begin{algorithmic}
        \FOR{$t=1$ to $N_{\textrm{MC}}$}
%        \STATE \emph{\scriptsize{\% Sampling the image means - see paragraph \ref{subsec:ima_mean}}}
%        \STATE Sample $\tilde{\bs{\mu}}_{\bfu}^{(t)}$ according to the conditional distribution \eqref{eq:imamean_post},
        \STATE \emph{\scriptsize{\% Sampling the image variances - see paragraph \ref{subsec:ima_var}}}
        \STATE Sample $\tilde{\bs{\Sigma}}_{\bfu}^{(t)}$ according to the conditional distribution \eqref{eq:imavar_post},
        \STATE \emph{\scriptsize{\% Sampling the high-resolved image - see paragraph \ref{subsec:ima}}}
        \STATE Sample $\sample{\bfu}{t}$ using an HMC algorithm detailed in Algo. \ref{algo:HMC} %in paragraph \ref{subsec:ima},
        \STATE \emph{\scriptsize{\% Sampling the noise variances - see paragraph \ref{subsec:noise_var}}}
        \FOR{$p=1$ to $P$}
				\FOR{$i=1$ to $\nbbandobs{p}$}
              	\STATE Sample ${\tilde{s}_{p,i}^{2(t)}}$ from the conditional distribution \eqref{eq:Varnoise_post},
            	\ENDFOR
            \ENDFOR
        \ENDFOR
    \end{algorithmic}
\end{algogo}
\end{figure}

%\subsection{Sampling the image mean $\meansub$ according to $f\left(\meansub|\Covsub,\bfu,\Vnoisevar,\allobs\right)$}
%\label{subsec:ima_mean} Combining \eqref{eq:Prior_mean} with
%\eqref{eq:prior_scene}, straightforward calculations lead to the
%following result
%\begin{equation}
%\label{eq:imamean_post} \meansub |\Covsub,\bfu,\Vnoisevar,\allobs
%\sim \calN \left(\imam{\meansub|\bfu },\imacovmat{\meansub|\bfu
%}\right)
%\end{equation}
%where $\imam{\meansub|\bfu } = \imacovmat{\meansub|\bfu}
%\left(\imacovmat{0}^{-1}\imam{0}+\Covsub^{-1}\sum\limits_{i=1}^{\nbrowima\nbcolima}
%\bsu_{i} \right)$ and $\imacovmat{\meansub |\bfu}=
%\left(\imacovmat{0}^{-1}+\nbrowima\nbcolima\Covsub^{-1}\right)^{-1}$.
%Note that this step requires the inversion of a $\wtm_{\lambda}
%\times \wtm_{\lambda}$ matrix, which can be achieved without any
%computational issue since $\wtm_{\lambda}$ is usually lower than
%$20$.

\subsection{Sampling $\Covsub$ according to $f\left(\Covsub|\bfu,\Vnoisevar,\allobs\right)$}
\label{subsec:ima_var} Standard computations yield the following
inverse-Wishart distribution as conditional distribution for the
covariance matrix $\Covsub$ (of the scene to be recovered)
\begin{equation}
\begin{split}
\label{eq:imavar_post}
&\Covsub |\bfu,\Vnoisevar,\allobs \sim \\
&\mathcal{W}^{-1}\left( \mathbf\Psi + \sum_{i=1}^{\nbrowima\nbcolima}(\bsu_i-\imam{\bsu_i})^T (\bsu_i-\imam{\bsu_i}),\nbrowima\nbcolima+\eta\right).\\
\end{split}
\end{equation}

\subsection{Sampling $\bfu$ according to $f\left(\bfu|\Covsub,\Vnoisevar,\allobs\right)$}
\label{subsec:ima} Choosing the conjugate distribution
\eqref{eq:prior_scene} as prior distribution for the projected
unknown image $\bfu$ leads to the following conditional posterior
distribution for $\bfu$
%\begin{equation}
%  \Vima|\meansub,\Covsub,\Vnoisevar,\allobs \sim \calN\left(\imam{\Vima|\allobs},\imacovmat{\Vima|\allobs}\right)
%\end{equation}
%with
%\begin{equation}
%\label{eq:param_normal}
%\begin{array}{ll}
%\imacovmat{\Vima|\allobs} &= \left[\imacovmat{\Vima}^{-1} + \sum_{\noobs=1}^{\nbobs}\frac{1}{\noisevar{\noobs}}\MATtrans{\noobs}\transp\MATtrans{\noobs}\right]^{-1}\\
%\imam{\Vima|\allobs}& =
%\imacovmat{\Vima|\allobs}\left[\sum_{\noobs=1}^{\nbobs}\frac{1}{\noisevar{\noobs}}\MATtrans{\noobs}\transp\Vobs{\noobs} +
%\imacovmat{\Vima}^{-1}\imam{\Vima} \right]
%\end{array}
%\end{equation}
\begin{equation}
  \bfu|\Covsub,\Vnoisevar,\allobs \sim \calN\left(\imam{\bfu|\allobs},\imacovmat{\bfu|\allobs}\right)
\end{equation}
with
\begin{equation}
\label{eq:param_normal}
\begin{array}{ll}
\imacovmat{\bfu|\allobs} &= \left[{\imacovall}^{-1} +
\sum_{\noobs=1}^{\nbobs}
\mathfrak{V}\MATtrans{\noobs}\transp {\bf\Lambda}_p^{-1} \MATtrans{\noobs}\mathfrak{V}^T\right]^{-1}\\
\imam{\bfu|\allobs}& =
\imacovmat{\bfu|\allobs}\left[\sum_{\noobs=1}^{\nbobs}\mathfrak{V}\MATtrans{\noobs}\transp {\bf\Lambda}_p^{-1}\Vobs{\noobs}
+ {\imacovall}^{-1}\imamall \right]
\end{array}
\end{equation}
%and
%\begin{equation*}
%\begin{split}
%\imacovmat{\Vima}= \mathfrak{V}^T  \imacovall \mathfrak{V}\\
%\imam{\Vima}= \mathfrak{V}^T \imamall.\\
%\end{split}
%\end{equation*}
%\imacovmat{\Vima}= \textrm{diag}[\underbrace{\mathbf{V}^T\Covsub\mathbf{V},\cdots,\mathbf{V}^T\Covsub\mathbf{V}}_{\nbrowima\nbcolima}] \\
%\imam{\Vima}= [\underbrace{\meansub^T\mathbf{V},\meansub^T\mathbf{V},\cdots,\meansub^T\mathbf{V}}_{\nbrowima\nbcolima}]^T.
Sampling directly according to this multivariate Gaussian
distribution requires the inversion of an $\wtM \times \wtM$ matrix,
which is impossible in most fusion problems. An alternative would
consist of sampling each element $u_{i}$ ($i=1,\ldots,\wtM$) of
$\bfu$ conditionally upon the others according to $f\left(u_{i}
|\bfu_{-i},\Vnoisevar,\Covsub,\allobs\right)$, where
$\bfu_{-i}$ is the vector $\bfu$ whose $i$th component has been
removed. However, this alternative would require to sample $\bfu$ by
using $\wtM$ Gibbs moves, which is time demanding and leads to poor
mixing properties.

%Indeed, this conditional distribution is a univariate
%normal distribution whose mean and variance can be easily derived by
%introducing partitioned covariance matrices and mean vectors \cite[p. 324]{Kay1988}.

The efficient strategy adopted in this work relies on a particular
MCMC method, called Hamiltonian Monte Carlo (HMC) method (sometimes
referred to as hybrid Monte Carlo method), which is considered to
generate vectors  $\bfu$ directly. %The goal of this
%generation is to exploit the fact that the observed vectors
%$\widetilde{\mathbf{x}}_i$ live in a subspace with reduced dimension
%and also to reduce the computational complexity of the sampling
%procedure.
More precisely, we consider the HMC algorithm initially proposed by
Duane \emph{et al.} for simulating the lattice field theory in
\cite{Duane1987}. As detailed in \cite{Neal1993}, this technique
allows mixing property of the sampler to be improved, especially in
a high-dimensional problem. It exploits the gradient of the
distribution to be sampled by introducing auxiliary ``momentum"
variables $\bfm\in\R^{\wtM}$. The joint distribution of the unknown
parameter vector $\mathbf{u}$ and the momentum is defined as
\begin{equation*}
f\left(\mathbf{u},\bfm|\Vnoisevar,\Covsub,\allobs\right)= f\left(\mathbf{u}|\Vnoisevar,\Covsub,\allobs\right)f\left(\bfm\right)
\end{equation*}
where $f\left(\bfm\right)$ is the normal probability density function (pdf) with zero mean and identity covariance matrix.
The Hamiltonian of the considered system is defined by taking the negative logarithm of the
posterior distribution
$f\left(\mathbf{u},\bfm|\Vnoisevar,\meansub,\Covsub,\allobs\right)$ to be sampled, i.e.,
\begin{equation}
\begin{array}{ll}
  H\left(\mathbf{u},\bfm\right) &= -\log f\left(\mathbf{u},\bfm|\Vnoisevar,\meansub,\Covsub,\allobs\right)\\
  &=U\left(\mathbf{u}\right) + K\left(\bfm\right)
\end{array}
\end{equation}
where $U\left(\mathbf{u}\right)$ is the potential energy function defined
by the negative logarithm of
$f\left(\mathbf{u}|\Vnoisevar,\Covsub,\allobs\right)$ and
$K\left(\bfm\right)$ is the corresponding kinetic energy
\begin{equation}
\begin{array}{ll}
 U\left(\mathbf{u}\right)&=-\log f\left(\mathbf{u}|\Vnoisevar,\Covsub,\allobs\right)\\
K\left(\bfm\right)&=\frac{1}{2}\bfm\transp\bfm.
\end{array}
\end{equation}
The parameter space where $\left(\mathbf{u},\bfm\right)$ lives is
explored following the scheme detailed in Algo~\ref{algo:HMC}. At
iteration $t$ of the Gibbs sampler, a so-called \emph{leap-frogging}
procedure composed of $N_\textrm{leapfrog}$ iterations is achieved
to propose a move from the current state
$\left\{\sample{\mathbf{u}}{t},\sample{\bfm}{t}\right\}$ to the
state
$\left\{\sample{\mathbf{u}}{\star},\sample{\bfm}{\star}\right\}$
with step size $\varepsilon$. This move is operated in
$\R^{\wtM}\times\R^{\wtM}$ in a direction given by the gradient of
the energy function
\begin{equation}
\label{eq:grad}
 \nabla_{\mathbf{u}}
                U\left(\mathbf{u}\right) = -\sum_{\noobs=1}^{\nbobs}
\mathfrak{V}
\MATtrans{\noobs}\transp {\bf\Lambda}_p^{-1}\left(\Vobs{\noobs}-\MATtrans{\noobs}\mathfrak{V}^T\mathbf{u}\right)
+ \imacovmat{\mathbf{u}}^{-1} (\mathbf{u}-\imamall).
\end{equation}
Then, the new state is accepted with probability $\rho_{t} =
\min\left\{1, A_{t}\right\}$ where
\begin{equation}
\label{eq:proba_HMCmove}
\begin{split}
  A_{t} &=
\frac{f\left(\sample{\mathbf{u}}{\star},\sample{\bfm}{\star}|\Vnoisevar,\Covsub,\allobs\right)}{f\left(\sample{\mathbf{u}}{t},\sample{\bfm}{t}|\Vnoisevar,\Covsub,\allobs\right)}\\
    &=
\exp\left[H\left(\sample{\mathbf{u}}{t},\sample{\bfm}{t}\right)-H\left(\sample{\mathbf{u}}{\star},\sample{\bfm}{\star}\right)\right].
\end{split}
\end{equation}

\begin{figure}[h!]
    \begin{algogo}{Hybrid Monte Carlo algorithm}
    \label{algo:HMC}
    \begin{algorithmic}
        \STATE \emph{\scriptsize{\% Momentum initialization}}
        \STATE Sample $\sample{\bfm}{\star} \sim \calN\left(\Vzeros{\wtM}, \Id{\wtM}\right)$,
        \STATE Set $\sample{\bfm}{t} \leftarrow \sample{\bfm}{\star}$,
        \STATE \emph{\scriptsize{\% Leapfrogging}}
         \FOR{$j=1~\text{to}~N_{\textrm{L}}$}
            \STATE Set $\sample{\bfm}{\star} \leftarrow \sample{\bfm}{\star} - \frac{\varepsilon}{2}\nabla_{\mathbf{u}}
                U\left(\sample{\mathbf{u}}{\star}\right)$,
            \STATE Set $\sample{\mathbf{u}}{\star} \leftarrow \sample{\mathbf{u}}{\star} + \varepsilon \sample{\bfm}{\star}$,
            \STATE Set $\sample{\bfm}{\star} \leftarrow \sample{\bfm}{\star} - \frac{\varepsilon}{2}\nabla_{\mathbf{u}}
                U\left(\sample{\mathbf{u}}{\star}\right)$,
         \ENDFOR
         \STATE \emph{\scriptsize{\% Accept/reject procedure, See \eqref{eq:proba_HMCmove}}}
        \STATE Sample $w \sim \calU\left([0,1]\right)$ ,
        \IF{$w < \rho_t$}
        \STATE $\sample{\mathbf{u}}{t+1}\leftarrow \sample{\mathbf{u}}{\star}$
        \ELSE
        \STATE $\sample{\mathbf{u}}{t+1}\leftarrow \sample{\mathbf{u}}{t}$
        \ENDIF
        \STATE Set $\sample{\Vima}{t+1}=\mathfrak{V}^T\sample{\mathbf{u}}{t+1}$
        \STATE Run Algo. \ref{algo:stepsize} to update stepsize
    \end{algorithmic}
\end{algogo}
\end{figure}

This accept/reject procedure ensures that the simulated vectors
$(\sample{\bfu}{t},\sample{\bfm}{t})$ are asymptotically distributed
according to the distribution of interest. The way the parameters
$\varepsilon$ and $N_\textrm{L}$ have been adjusted will be detailed
in Section \ref{sec:simulation}.
%Again, note that after generating a
%sample $\sample{\bfu}{t}$ in $\mathbb{R}^{\wtM}$, the corresponding
%sample $\sample{\bfx}{t}$ of the image $\Vima$ to be recovered can
%be obtained from the transformation
%\begin{equation}
%\label{eq:trans_low2high} \sample{\bfx}{t}= \mathfrak{V}^T
%\sample{\bfu}{t}.
%\end{equation}

To sample according to a high-dimension Gaussian distribution such
as $f\left(\bfu|\Covsub,\Vnoisevar,\allobs\right)$, one
might think of using other simulation techniques such as the method
proposed in \cite{Zhang2012Generative} to solve super resolution
problems. Similarly, Orieux \emph{et al.} have proposed a
perturbation approach to sample high-dimensional Gaussian
distributions for general linear inverse problems \cite{Orieux2012}.
However, these techniques rely on additional optimization schemes
included within the Monte Carlo algorithm, which implies that the
generated samples are only approximately distributed according to
the target distribution. Conversely, the HMC strategy proposed here
ensures asymptotic convergence of the generated samples to the
posterior distribution. Moreover, the HMC method is very flexible
and can be easily extended to handle non-Gaussian posterior
distributions contrary to the methods investigated in
\cite{Zhang2012Generative,Orieux2012}.

\subsection{Sampling $\Vnoisevar$ according to $f\left(\Vnoisevar|\bfu,\Covsub,\allobs\right)$}
\label{subsec:noise_var}
The conditional pdf of the noise variance
$\noisevar{\noobs,i}$ ($i=1,\ldots,\nbbandobs{\noobs}, \noobs=1,\ldots,\nbobs$)
is
\begin{equation}
\begin{split}
   & f\left(\noisevar{\noobs,i}|\bfu,\Covsub,\allobs\right) \propto \\
& \left(\frac{1}{\noisevar{\noobs,i}}\right)^{\frac{\nbrowobs{\noobs}\nbcolobs{\noobs}}{2}+1}
\exp\left(-\frac{\norm{(\Vobs{\noobs}-\MATtrans{\noobs}\mathfrak{V}^T\bfu)_i}^2}{2\noisevar{\noobs,i}}\right)
\end{split}
\end{equation}
where $(\Vobs{\noobs}-\MATtrans{\noobs}\mathfrak{V}^T\bfu)_i$ contains the elements of the $i$th band.
Generating samples $\noisevar{\noobs,i}$ distributed according to
$f\left(\noisevar{\noobs,i}|\bfu,\Covsub,\allobs\right)$ is
classically achieved by drawing samples from the following
inverse-gamma distribution
\begin{equation}
\label{eq:Varnoise_post}
  \noisevar{\noobs,i} |\bfu,\allobs \sim \calI\calG\left(\frac{\nbrowobs{\noobs}\nbcolobs{\noobs}}{2},\frac{\norm{(\Vobs{\noobs}-\MATtrans{\noobs}\mathfrak{V}^T\bfu)_i}^2}{2}\right).
\end{equation}

In practice, if the noise variances are known a prior, we simply assign the noise variances
to be known values and remove the sampling of the noise variances.

\subsection{Complexity Analysis}
The MCMC method can be computationally costly compared with optimization methods \cite{David2012}.
The complexity of the proposed Gibbs sampler is mainly due to the Hamiltonian Monte Carlo method.
The complexity of the Hamiltonian MCMC method is
$\mathcal {O}((\widetilde{\nbbandima})^3)+ \mathcal {O}((\widetilde{\nbbandima}\nbrowima\nbcolima)^2)$,
which is highly expensive as $\nbbandima$ increases. Generally the number of pixels $\nbrowima\nbcolima$
cannot be reduced significantly. Thus, projecting the high-dimensional $\nbbandima \times 1$ vectors to
a low-dimension space to form $\widetilde{\nbbandima} \times 1$ vectors decreases the complexity while
keeping most important information.
%As there exists matrix product and inversion of matrix $\Covsub$ to evaluate the gradient in \eqref{eq:grad},
\section{Simulation Results}
\label{sec:simulation}

This section studies the performance of the proposed Bayesian fusion
algorithm. The reference image, considered here as the high spatial
and high spectral image, is an hyperspectral image acquired over
Moffett field, CA, in 1994 by the JPL/NASA airborne visible/infrared
imaging spectrometer (AVIRIS) \cite{Green1998imaging}. This image
was initially composed of $224$ bands that have been
reduced to $177$ bands ($\nbbandima=\nbbandobs{1}=177$) after removing the water vapor absorption bands.

\subsection{Fusion of HS and MS images}
\label{subsec:fuse_HS_MS}

\begin{figure}
\centering
    \includegraphics[width=0.5\textwidth]{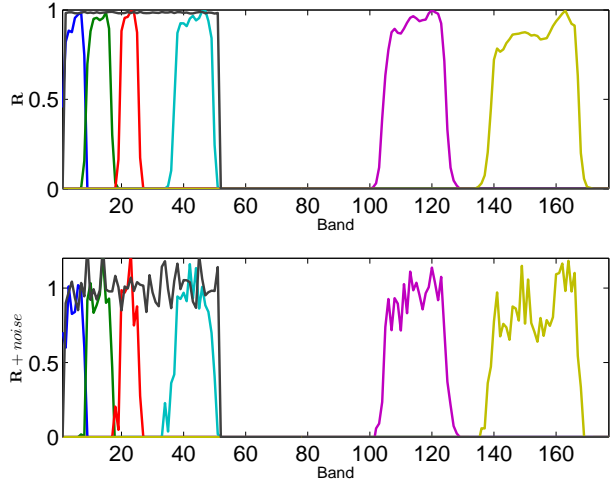}%}
    \caption{LANDSAT spectral responses. (Top) without noise. (Bottom) with an additive Gaussian noise with FSNR = 8dB.}
    \label{fig:F2}
\end{figure}

We propose to reconstruct the reference HS image from two
lower resolved images. First, a high-spectral low-spatial resolution
image $\Vobs{1}$, denoted as HS image, has been generated by
applying a $5 \times 5$ averaging filter on each band of the
reference image. Besides, an MS image $\Vobs{2}$ is obtained by
successively averaging the adjacent bands according to realistic
spectral responses. More precisely, the reference image is filtered
using the LANDSAT-like spectral responses depicted in the top of Fig. \ref{fig:F2},
to obtain a $7$-band ($\nbbandobs{2}=7$) MS image. Note here that the observation
models $\MATtrans{1}$ and $\MATtrans{2}$ corresponding to the HS and
MS images are perfectly known. In addition to the blurring and
spectral mixing, the HS and MS images have been both contaminated by
zero-mean additive Gaussian noise. The noise power $\noisevar{\noobs,i}$
depends on the signal to noise ratio SNR$_{p,i}$ $\left(i=1,\cdots,\nbbandobs{p}, p=1,2\right)$ defined by
$\textrm{SNR}_{p,i}=10\log_{10} \left(\frac{\|\left(\MATtrans{\noobs}{\Vima}\right)_i\|_F^2}{\nbrowobs{p}\nbcolobs{p}\noisevar{\noobs,i}}\right)$,
where $\|{.}\|_F$ is the Frobenius norm.

\begin{figure}[htb]
\centering
    \subfigure{
    \label{fig:subfig:HS}
    \includegraphics[width=0.12\textwidth]{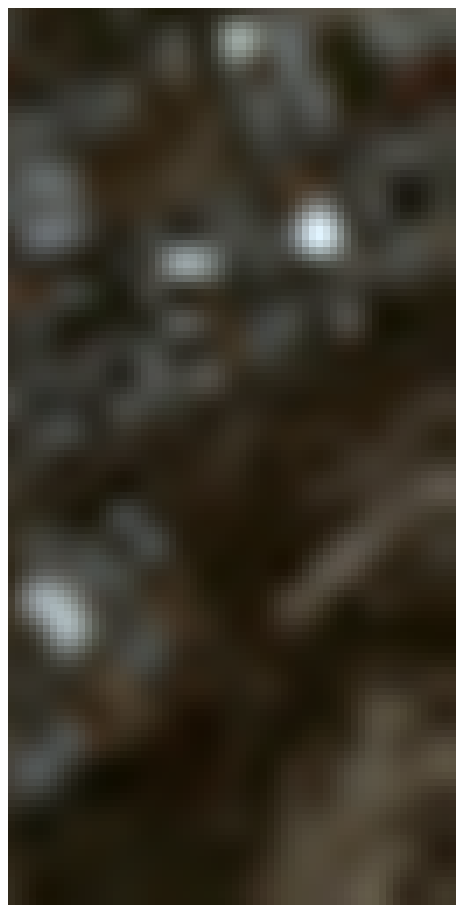}}
    \hspace{0in}
    \subfigure{
    \label{fig:subfig:MS}
    \includegraphics[width=0.12\textwidth]{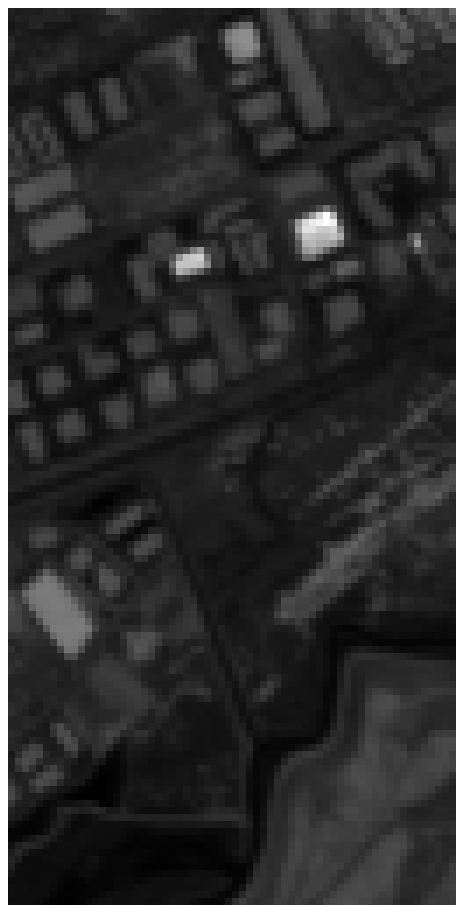}}
    \hspace{0in}
    \subfigure{
    \label{fig:subfig:Hardie}
    \includegraphics[width=0.12\textwidth]{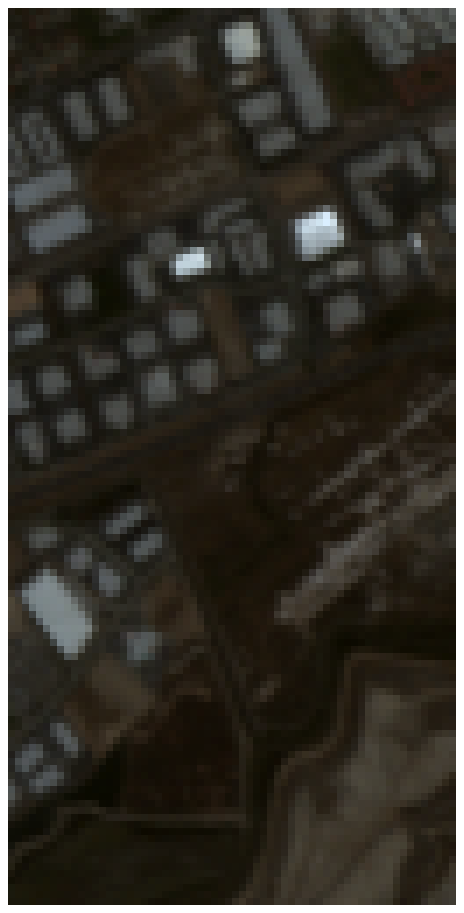}}\\
    \hspace{0in}
    \subfigure{
    \label{fig:subfig:Zhang}
    \includegraphics[width=0.12\textwidth]{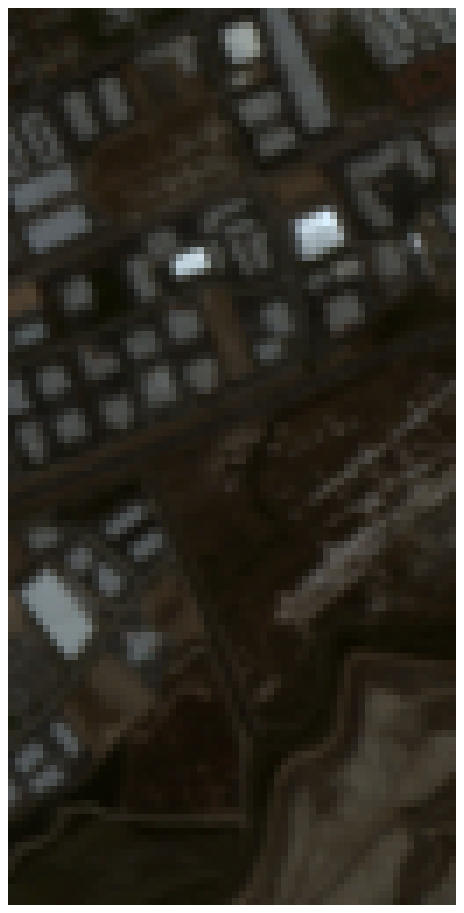}}
    \hspace{0in}
    \subfigure{
    \label{fig:subfig:HMC}
    \includegraphics[width=0.12\textwidth]{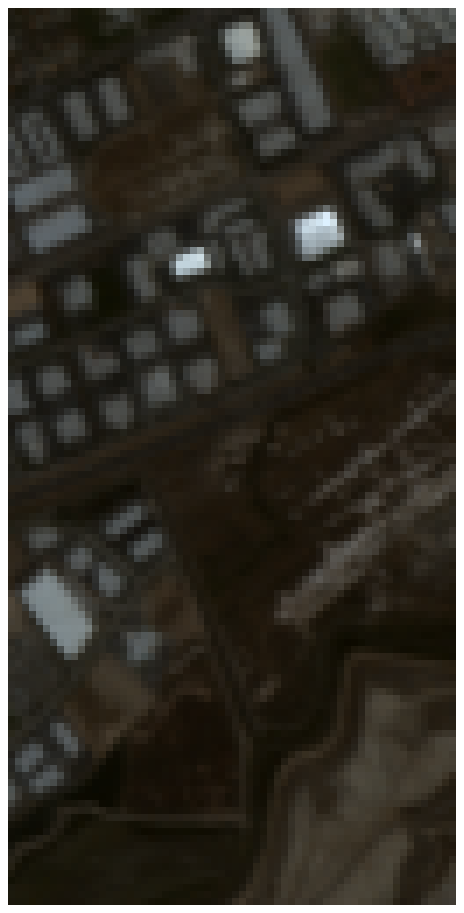}}
    \hspace{0in}
    \subfigure{
    \label{fig:subfig:Ref}
    \includegraphics[width=0.12\textwidth]{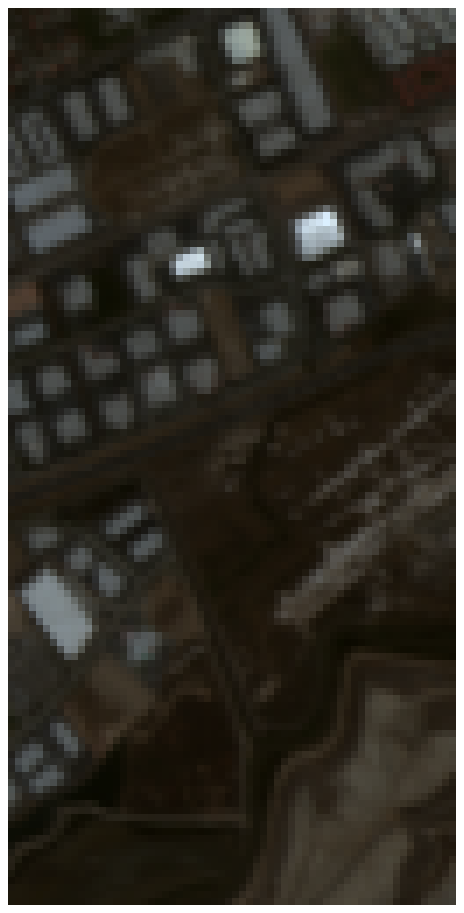}}
    \hspace{0in}
    \caption{AVIRIS dataset: (Top left) HS Image. (Top middle) MS Image. (Top right) MAP \cite{Hardie2004}.
    (Bottom left) Wavelet MAP \cite{Zhang2009}. (Bottom middle) Hamiltonian MCMC. (Bottom right) Reference.}
    \label{fig:O_MS_HS}
\end{figure}

Our simulations have been conducted with SNR$_{1,\cdot}=35$dB for the first 127 bands and
SNR$_{1,\cdot}=30$dB for the remaining 50 bands of the HS image. For the MS image, SNR$_{2,\cdot}$
is 30dB for all bands. A composite color image, formed by selecting the red, green and blue bands
of the high-spatial resolution HS image (the reference image) is shown in the right bottom of Fig.
\ref{fig:O_MS_HS}. The noise-contaminated HS and MS images are
depicted in the top left and top right of Fig. \ref{fig:O_MS_HS}.

\subsubsection{Subspace learning}
\label{subsubsec:subspace}
\begin{figure}[htb]
\centering
\includegraphics[width=0.3\textwidth]{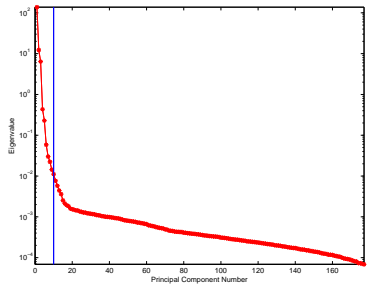}
\caption{Eigenvalues of $\bs{\Upsilon}$ for the HS image.}
\label{fig:Eigen_value}
\end{figure}

To learn the matrix $\mathbf{V}$ in \eqref{eq:subspace}, we propose to use the principal component analysis (PCA)
which is a classical dimensionality reduction technique used in HS imagery. %\cite{Farrell2005}.
As in paragraph \ref{subsubsec:scene_prior}, the vectorized HS image
$\Vobs{1}$ can be written as
 $\Vobs{1} = \left[\bsz_{1,1}^T,\bsz_{1,2}^T,\cdots,\bsz_{1,\nbrowobs{1}\nbcolobs{1}}^T\right]^T$,
where
$\bsz_{1,i}=\left[z_{1,i,1},z_{1,i,2},\cdots,z_{1,i,\nbbandobs{1}}\right]^T$.
Then, the sample covariance matrix of the HS image $\Vobs{1}$
%defined as
%\begin{equation}
%\label{eq:est_var_band}
%\begin{array}{ll}
%\boldsymbol{\Lambda}_1 & =
%\frac{1}{\nbrowobs{1}\nbcolobs{1}-1}\sum\limits_{i=1}^{\nbrowobs{1}\nbcolobs{1}}
%(\bsz_{1,i}-\bar{\bsz}_1)(\bsz_{1,i}-\bar{\bsz}_1)^T\\
%\bar{\bsz}_1& =
%\frac{1}{\nbrowobs{1}\nbcolobs{1}}\sum\limits_{i=1}^{\nbrowobs{1}\nbcolobs{1}}\bsz_{1,i}
%\end{array}
%\end{equation}
is diagonalized leading to
\begin{equation}
\mathbf{W}^{T} \bs{\Upsilon} \mathbf{W} = \mathbf{D}
\end{equation}
where $\mathbf{W}$ is an $\nbbandima \times \nbbandima$ orthogonal
matrix ($\mathbf{W}^T=\mathbf{W}^{-1}$) and $\mathbf{D}$ is a
diagonal matrix whose diagonal elements are the ordered eigenvalues
of $\bs{\Upsilon}$ denoted as $d_1 \ge d_2 \ge ... \ge
d_{\nbbandima}$. The dimension of the projection subspace
$\wtm_\lambda$ is defined as the minimum integer satisfying the
condition ${\sum_{i=1}^{\wtm_{\lambda}}d_i}/{\sum_{i=1}^{\nbbandima}d_i}\ge 0.99$.
%\begin{equation} \label{m_lambda}
%\end{equation}
The matrix $\mathbf{V}$ is then constructed as the eigenvectors
associated with the $\wtm_\lambda$ largest eigenvalues of
$\bs{\Upsilon}$.  As an illustration, the eigenvalues of
the sample covariance matrix $\bs{\Upsilon}$ for the
Moffett field image are displayed in Fig. \ref{fig:Eigen_value}. For
this example, the $\wtm_{\lambda}= 10$ eigenvectors contain $99.93 \%$
of the information.
%(i.e., $\wtm_{\lambda}=18$ is the smallest
%integer satisfying \eqref{m_lambda}).

\subsubsection{Hyper-hyperparameters selection}

In our experiments, fixed hyper-hyperparameters have been chosen as
follows: $ \mathbf{\Psi} = \Id{\wtm_{\lambda}},
\eta  = \wtm_{\lambda} +3.$

These choices can be motivated by the following arguments:
\begin{itemize}
\item The identity matrix assigned to $\mathbf{\Psi}$ ensures a non-informative prior.
\item Setting the inverse gamma parameters to $\eta = \wtm_{\lambda} +3$ also
leads to a non-informative prior \cite{Punskaya2002}.
\item Note that parameter $\nu$ disappears when the joint posterior is
integrated out with respect to parameter $\gamma$.
\end{itemize}

\subsection{Stepsize and Leapfrog Steps}
\label{subsubsec:stepsize} The performance of the HMC method is
mainly governed by the stepsize $\varepsilon$ and the number of
leapfrog steps $N_\textrm{L}$. As pointed out in \cite{Neal2010}, a
too large stepsize will result in a very low acceptance rate and a
too small stepsize yields high computational complexity. In order to
adjust the stepsize parameter $\varepsilon$, we propose to monitor
the statistical acceptance ratio $\hat{\rho}_t$ defined as $\hat{\rho}_t = \frac{N_\textrm{a,t}}{N_\textrm{W}}$
%\begin{equation}
%\label{eq:Accept}
%\end{equation}
where $N_\textrm{W}$ is the length of the counting window
(in our experiment, the counting window at time $t$
contains the vectors
$\sample{\Vima}{t-N_\textrm{W}+1},\sample{\Vima}{t-N_\textrm{W}},\cdots,\sample{\Vima}{t}$
with $N_\textrm{W} = 50$) and  $N_\textrm{a,t}$ is the number of
accepted samples in this window at time $t$. As explained in \cite{Roberts2007},
the adaptive tuning should adapt less and less as the algorithm
proceeds to guarantee that the generated samples form a
stationary Markov chain. In the proposed implementation,  the
parameter $\varepsilon$ is adjusted as in Algo. \ref{algo:stepsize}.
The thresholds have been fixed to $(\alpha_d,\alpha_u)=
(0.3, 0.9)$ and the scale parameters are
$(\beta_d,\beta_u)=(1.1,0.9)$ (these parameters were adjusted by
cross-validation). Note that the initial value of $\varepsilon$
should not be too large to `blow up' the leapfrog trajectory
\cite{Neal2010}. Generally, the stepsize converges after some
iterations of Algo. \ref{algo:stepsize}.

\begin{figure}[h!]
\begin{algogo}{Adjusting Stepsize}
\label{algo:stepsize}
\begin{algorithmic}
    \STATE Update $\hat{\rho}_t$ with $N_\textrm{a,t}$ : $\hat{\rho}_t = \frac{N_\textrm{a,t}}{N_\textrm{W}}$
    \STATE \emph{\scriptsize{\% Burn-in ($t \leq N_\textrm{MC}$):}}
    \IF{$\hat{\rho}_t > \alpha_u$}
    \STATE Set $\varepsilon = \beta_u \varepsilon$
    \ELSIF{$\hat{\rho}_t < \alpha_d$}
        \STATE Set $\varepsilon = \beta_d \varepsilon$
    \ENDIF
    \STATE \emph{\scriptsize{\% After Burn in ($t>N_\textrm{MC}$):}}
    \IF{$\hat{\rho}_t > \alpha_u$}
    \STATE Set $\varepsilon = \left[1-(1-\beta_u)\textrm{exp}(-0.01\times(t-N_\textrm{bi})\right)]\varepsilon$, \\
    \ELSIF{$\hat{\rho}_t < \alpha_d$}
        \STATE Set $\varepsilon = \left[1-(1-\beta_d)\textrm{exp}(-0.01\times(t-N_\textrm{bi})\right)]\varepsilon$, \\
    \ENDIF \\
    $(t= N_\textrm{bi}+1,\cdots,N_\textrm{MC})$

\end{algorithmic}
\end{algogo}
\end{figure}

Regarding the number of leapfrogs, setting the trajectory length $N_\textrm{L}$ by
trial and error is necessary \cite{Neal2010}. To avoid the potential resonance, $N_\textrm{L}$ is randomly chosen from a uniform distribution from $N_\textrm{min}$ to $N_\textrm{max}$.
After some preliminary runs and tests, $N_\textrm{min} =50$ and $N_\textrm{max} = 55$ have been selected.

\subsection{Evaluation of the Fusion Quality}
To evaluate the quality of the proposed fusion strategy, different image quality measures can be investigated.
Referring to \cite{Zhang2009}, we propose to use RSNR, SAM, UIQI, ERGAS and DD as defined below.
\paragraph{RSNR}
The reconstruction SNR (RSNR) is related to the difference between the actual and fused images
\begin{equation}
\textrm{RSNR}(\Vima,\hat{\Vima})=10\log_{10} \left(\frac{\|\Vima\|^2}{\|\Vima-\hat{\Vima}\|_2^2}\right).
\label{eq:SNR}
\end{equation}
The larger RSNR, the better the fusion quality and vice versa.
\paragraph{SAM}
The spectral angle mapper (SAM) measures the spectral distortion between the actual and estimated images.
The SAM of two spectral vectors $\bsx_n$ and $\hat{\bsx}_n$ is defined as
\begin{equation}
\textrm{SAM}(\bsx_n,\hat{\bsx}_n)=\textrm{arccos} \left(\frac{\langle\bsx_n,\hat{\bsx}_n\rangle}{ \|\bsx_n\|_2\|\hat{\bsx}_n\|_2}\right).
\label{eq:SAM}
\end{equation}
The average SAM is finally obtained by averaging the SAMs of all image pixels.
Note that SAM value is expressed in radians and thus belongs to $[-\frac{\pi}{2},\frac{\pi}{2}]$.
The smaller the absolute value of SAM, the less important the spectral distortion.

\paragraph{UIQI}
The universal image quality index (UIQI) was proposed in \cite{Wang2002} for evaluating the similarity between two single band images. It is related to the correlation, luminance distortion and contrast distortion of the estimated image to the reference image. The UIQI between $\bfa=[a_1,a_2,\cdots,a_N]$ and $\hat{\bfa}=[\hat{a}_1,\hat{a}_2,\cdots,\hat{a}_N]$ is defined as
\begin{equation}
\textrm{UIQI}(\bfa,\hat{\bfa}) =
\frac{4\sigma_{a \hat{a}}^2 \mu_{a} \mu_{\hat{a}}}{(\sigma_{a}^2+\sigma_{\hat{a}}^2)(\mu_{a}^2+\mu_{\hat{a}}^2)}
\label{eq:UIQI}
\end{equation}
where
%\begin{eqnarray*}
%\mu_{a} =\frac{1}{N}\sum_{i=1}^{N} a_i \\
%\mu_{\hat{a}}=\frac{1}{N}\sum_{i=1}^{N}\hat{a}_i \\
%\sigma_{a}^2=\frac{1}{N-1}\sum_{i=1}^{N}(a_i-\mu_{a})^2 \\
%\sigma_{\hat{a}}^2=\frac{1}{N-1}\sum_{i=1}^{N}(\hat{a}_i-\mu_{\hat{a}})^2 \\
%\sigma_{a \hat{a}}=\frac{1}{N-1}\sum_{i=1}^{N}(a_i-\mu_{a})(\hat{a}_i-\mu_{\hat{a}})
%\end{eqnarray*}
%}
$\left(\mu_{a},\mu_{\hat{a}},\sigma_{a}^2,\sigma_{\hat{a}}^2\right)$
are the sample means and variances of $a$ and $\hat{a}$, and
$\sigma_{a \hat{a}}^2$ is the sample covariance of
$\left(a,\hat{a}\right)$. The range of UIQI is $[-1,1]$ and UIQI$=1$
when $\bfa=\hat{\bfa}$. For multi-band image, the UIQI is obtained
band-by-band and averaged over all bands.
\paragraph{ERGAS}
The relative dimensionless global error in synthesis (ERGAS) calculates the amount of spectral
distortion in the image \cite{Wald2000}. This measure of fusion quality is defined as
\begin{equation}
\textrm{ERGAS}=100 \times \frac{1}{d^2} \sqrt{\frac{1}{\nbbandima}\sum_{i=1}^{\nbbandima}\left(\frac{\textrm{RMSE}(i)}{\mu_i}\right)}
\end{equation}
where $1/d^2$ is the ratio between the pixel sizes of the MS and HS images, $\mu_i$ is the mean of the $i$th band of the HS image, and $\nbbandima$ is the number of HS bands. The smaller ERGAS, the
smaller the spectral distortion.

\paragraph{DD}
The degree of distortion (DD) between two images $\MATima$ and $\hat{\MATima}$ is defined as
\begin{equation}
\textrm{DD}(\MATima,\hat{\MATima})=\frac{1}{\nbpixima}\|\textrm{vec}(\MATima)-\textrm{vec}(\hat{\MATima})\|_1.
\end{equation}
The smaller DD, the better the fusion.

\subsection{Comparison with other Bayesian models}
The Bayesian model proposed here differs from previous Bayesian models \cite{Hardie2004,Zhang2009} in three-fold.
First, in addition to the target image $\Vima$, the hierarchical Bayesian model allows the distributions of
the noise variances $\Vnoisevar$ and the hyperparameter $\Covsub$ to be inferred.
The hierarchical inference structure makes this Bayesian model more general and flexible. Second,
the covariance matrix $\Covsub$ is assumed to be block diagonal,
which allows us to exploit the correlations between spectral bands. Third, the proposed method
takes advantage of the relation
between the multispectral image and the target image by introducing a forward model $\MATtrans{2}$.
This paragraph compares the proposed Bayesian fusion method with
these two state-of-the-art fusion algorithms \cite{Hardie2004}
\cite{Zhang2009} for HS+MS fusion. The
MMSE estimator of the image using the proposed Bayesian method is
obtained from \eqref{eq:MMSE_X}. In this simulation, $N_{\textrm{MC}}=500$
and $N_{\textrm{bi}}=500$. The fusion results obtained with
different algorithms are depicted in Fig.
\ref{fig:O_MS_HS}. Graphically, the proposed algorithm performs
competitively with the state-of-the-art methods. This result is
confirmed quantitatively in Table \ref{tb:quality} which shows the
RSNR, UIQI, SAM, ERGAS and DD for the three methods.
%To illustrate the performance of the HMC method in different noise scenarios, Fig.
%\ref{fig:Performance_SNR2} compares the fusion quality measures RE,
%SAM and UIQI for different fusion methods.
It can be seen that the HMC method provides slightly better results
in terms of image restoration than the other methods. However, the proposed method
allows the image covariance matrix and the noise variances to be estimated. The samples
generated by the MCMC method can also be used to compute confidence intervals for the estimators
(e.g., see error bars in Fig. \ref{fig:Var_Noise}).
%(at the price of a higher computational complexity).

\begin{table}
\centering
\caption{Performance of different fusion methods in terms of: RSNR (\lowercase{dB}), UIQI, SAM (deg), ERGAS and DD($\times 10^{-2}$) (AVIRIS dataset).}
\begin{tabular}{c|c|c|c|c|c|c}
\hline
Methods & RSNR & UIQI & SAM  &ERGAS & DD  & Time\\
\hline
MAP      & 23.33  & 0.9913 & 5.05 &4.21 & 4.87 &  $\bs{1.6}$\\
Wavelet  & 25.53  & 0.9956 & 3.98 &3.95 & 3.89 & 31\\
Proposed & $\bs{26.74}$  & $\bs{0.9966}$ & $\bs{3.40}$ & $\bs{3.77}$ & $\bs{3.33}$ & 530\\
\hline
\end{tabular}
\label{tb:quality}
\end{table}

\subsection{Estimation of the noise variances}
\begin{figure}
\centering
\subfigure{
\includegraphics[width=0.5\textwidth]{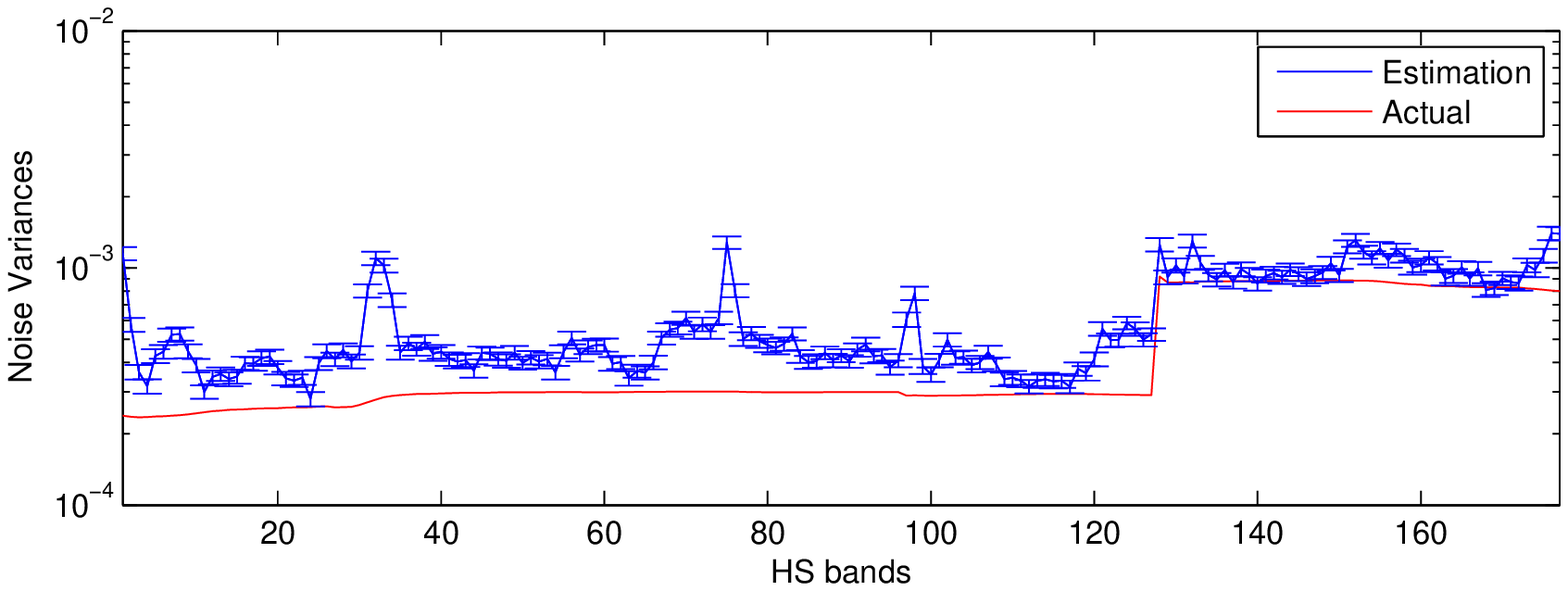}}
\subfigure{
\includegraphics[width=0.5\textwidth]{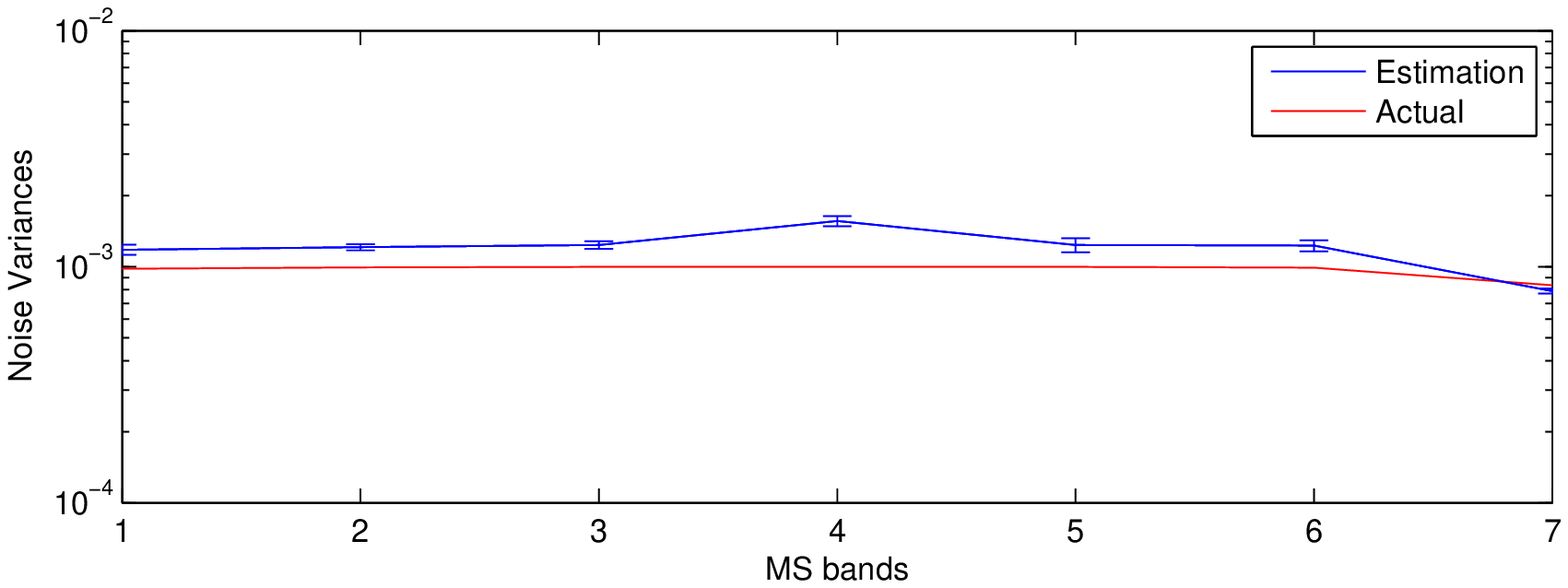}}
\caption{Noise variances and their MMSE estimates. (Top) HS image. (Bottom) MS image.}
\label{fig:Var_Noise}
\end{figure}

The proposed Bayesian method allows noise variances $\noisevar{p,i}$
$(i=1,\cdots,\nbbandobs{p},p=1,\cdots,P)$ to be estimated from the samples generated by the
Gibbs sampler. The MMSE estimators of $\noisevar{1,(\cdot)}$ and $\noisevar{2,(\cdot)}$
are illustrated in Fig. \ref{fig:Var_Noise}. Graphically, the estimations
can track the variations of the noise powers within tolerable discrepancy.

\subsection{Robustness with respect to the knowledge of $\MATtrans{2}$}
\label{subsec:F2}
\begin{figure}
\centering
\includegraphics[width=0.5\textwidth]{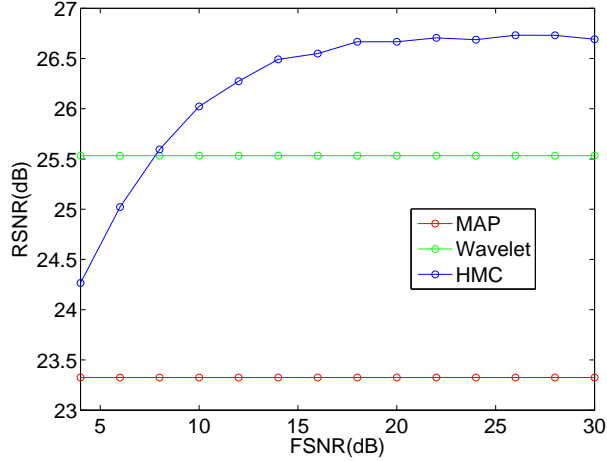}
\caption{Reconstruction errors of the different fusion methods versus FSNR.}
\label{fig:Test_F2}
\end{figure}

The sampling algorithm summarized in Algo. \eqref{algo:HMC} requires the knowledge of the spectral response $\MATtrans{2}$. However, this knowledge can be partially known in some practical applications. As the spectral response is the same for each vector $\bsx_i$ $(i=1,\cdots,\nbrowima\nbcolima)$, $\MATtrans{2}$ can be constructed from the matrix $f_2$ of size $\nbbandobs{2}\times\nbbandima$ (i.e., $7\times172$) as follows
\begin{equation}
\label{eq:F2_big}
\MATtrans{2}= \textrm{diag}[\underbrace{f_2 \cdots f_2}_{\nbrowima \nbcolima}].
\end{equation}
This paragraph is devoted to testing the robustness of the proposed
algorithm to the imperfect knowledge of $f_2$. In order to analyze
this robustness, a zero-mean white Gaussian error has been added to
any non-zero component of $f_2$ as shown in the bottom of Fig. \ref{fig:F2}.
Of course, the level of uncertainty regarding $f_2$ is controlled by
the variance of the error denoted as $\sigma_2^2$. The corresponding
FSNR is defined to adjust the knowledge of $f_2$:
\begin{equation}
\label{eq:PSNR}
\textrm{FSNR}=10\log_{10} \left(\frac{\|f_2\|_F^2}{\nbbandima\nbbandobs{2}\noisevar{2}}\right).
\end{equation}
The larger FSNR, the more knowledge we have about $f_2$. The
RSNRs between the reference and estimated images are
displayed in Fig. \ref{fig:Test_F2} as a function of FSNR. Obviously, the performance of the proposed Bayesian
fusion algorithm decreases as the uncertainty about $f_2$ increases.
However, as long as the FSNR is above $8$dB, the performance of
the proposed method always outperforms the MAP and wavelet-based MAP
methods. Thus, the proposed method is quite robust with respect to
the imperfect knowledge of $f_2$.

\subsection{Test on additional dataset}
\label{subsec:Pavia}
\begin{figure}[htb]
\centering
    \subfigure{
    \label{fig:subfig:HS}
    \includegraphics[width=0.12\textwidth]{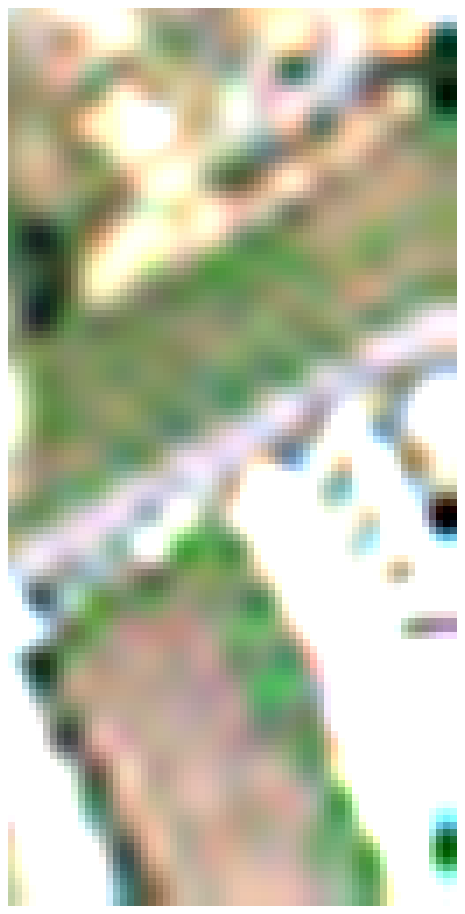}}
    \hspace{0in}
    \subfigure{
    \label{fig:subfig:MS}
    \includegraphics[width=0.12\textwidth]{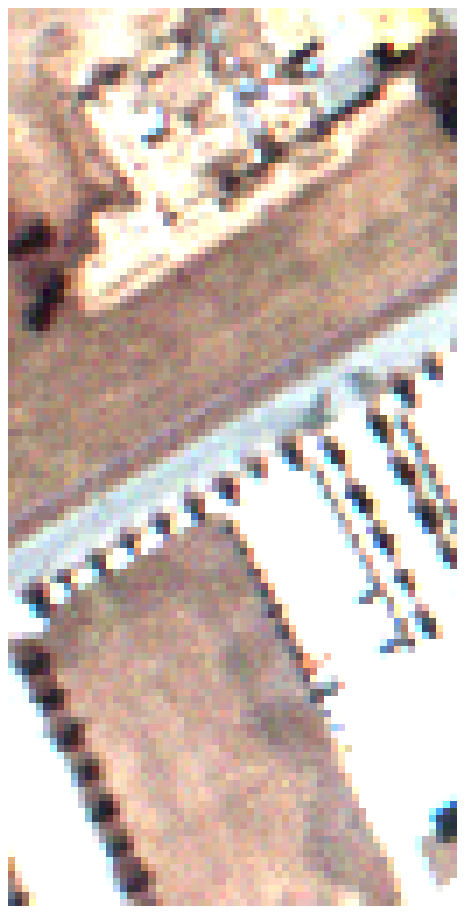}}
    \hspace{0in}
    \subfigure{
    \label{fig:subfig:Hardie}
    \includegraphics[width=0.12\textwidth]{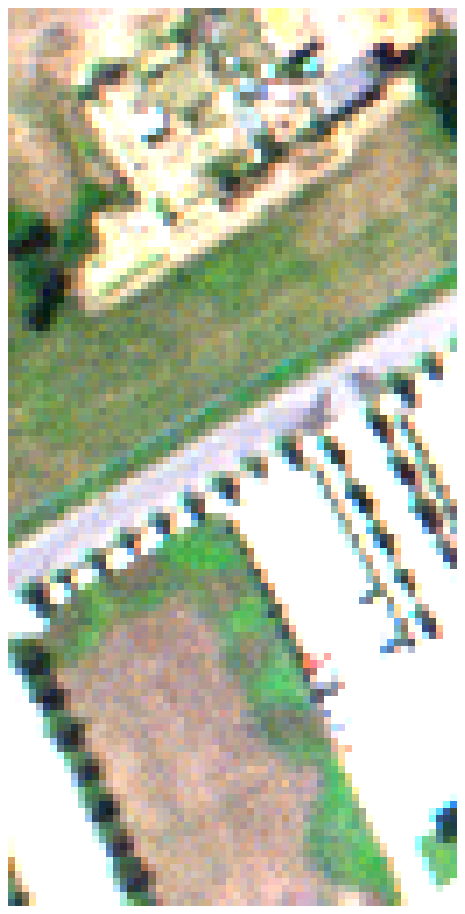}}\\
    \hspace{0in}
    \subfigure{
    \label{fig:subfig:Zhang}
    \includegraphics[width=0.12\textwidth]{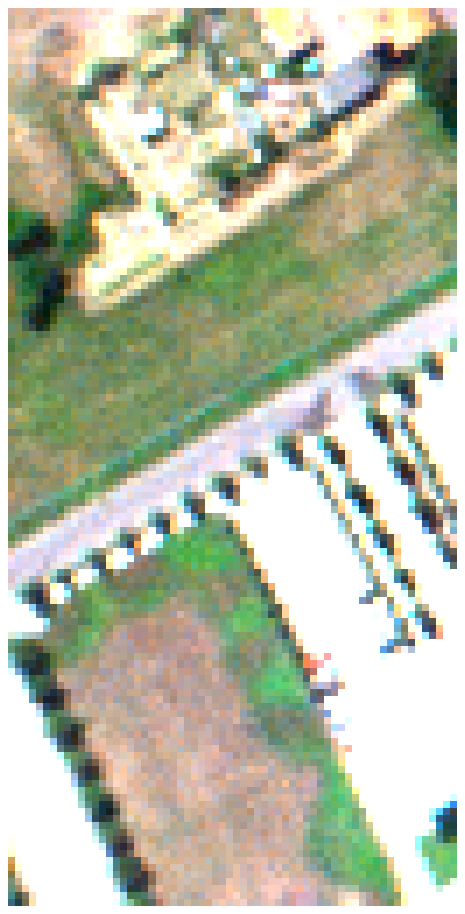}}
    \hspace{0in}
    \subfigure{
    \label{fig:subfig:HMC}
    \includegraphics[width=0.12\textwidth]{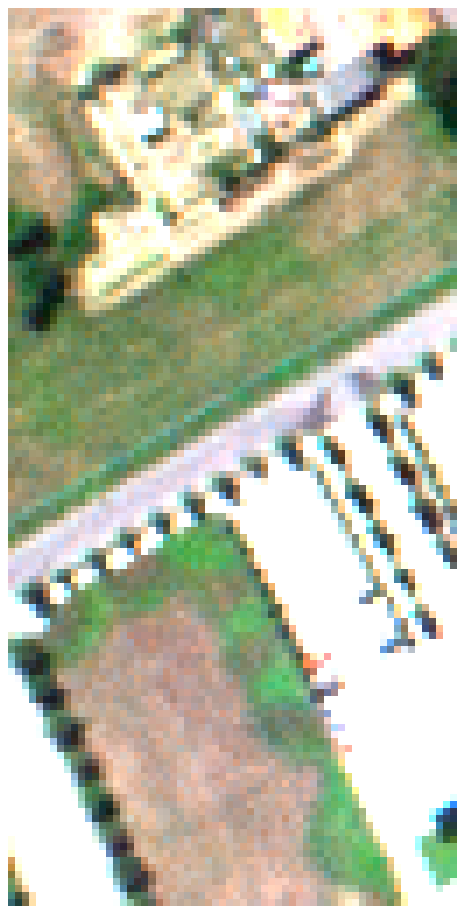}}
    \hspace{0in}
    \subfigure{
    \label{fig:subfig:Ref}
    \includegraphics[width=0.12\textwidth]{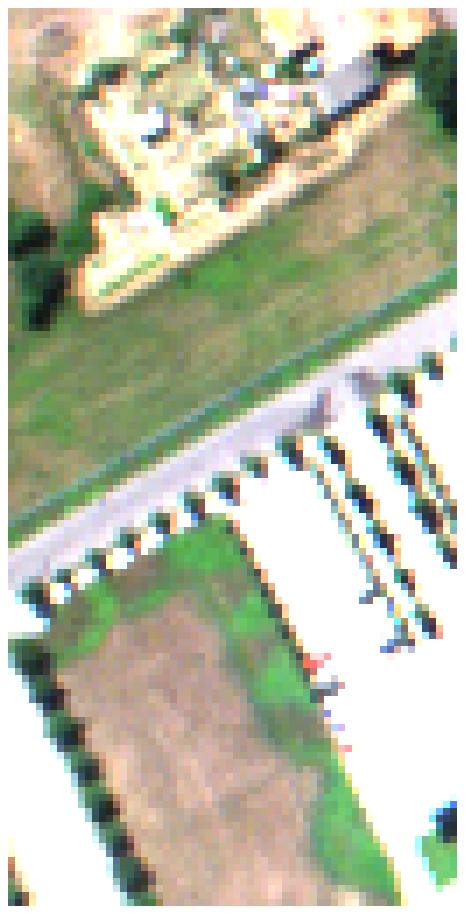}}
    \hspace{0in}
    \caption{ROSIS dataset: (Top left) HS Image. (Top middle) MS Image. (Top right) MAP \cite{Hardie2004}.
    (Bottom left) Wavelet MAP \cite{Zhang2009}. (Bottom middle) Hamiltonian MCMC. (Bottom right) Reference.}
    \label{fig:Pavia}
\end{figure}

\begin{figure}
\centering
\subfigure{
\includegraphics[width=0.5\textwidth]{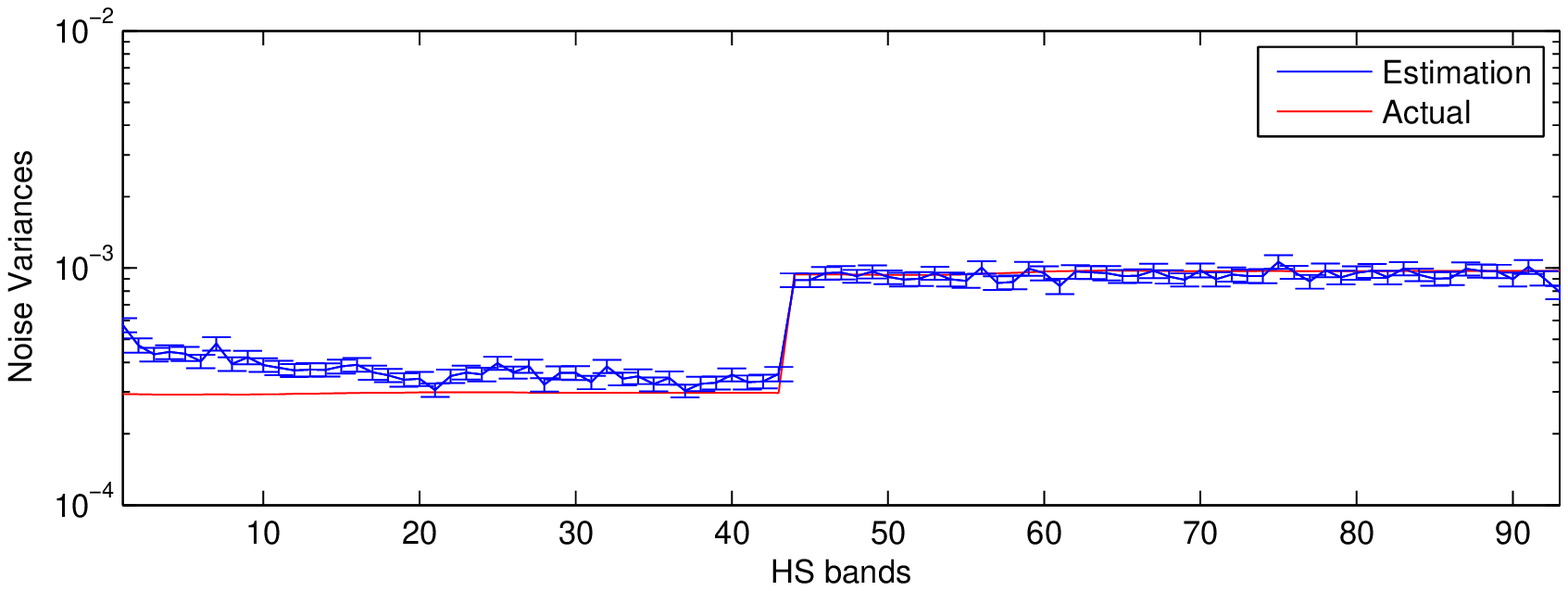}}
\subfigure{
\includegraphics[width=0.5\textwidth]{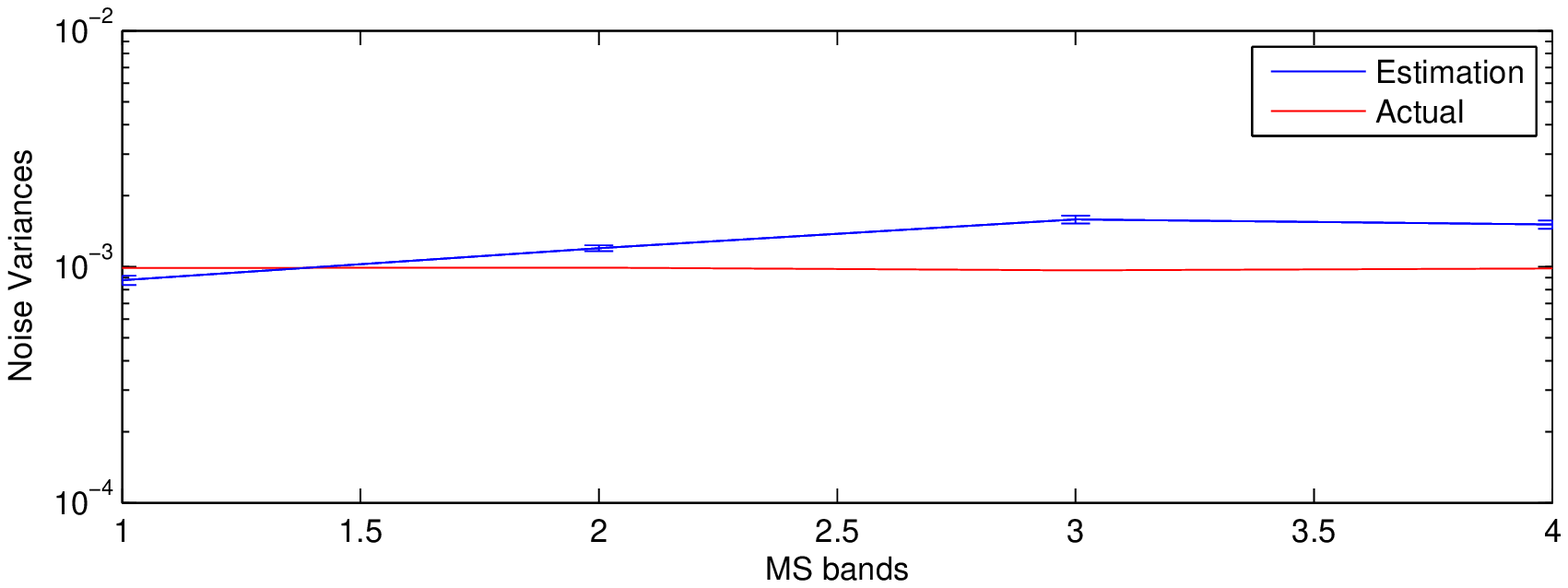}}
\caption{Noise variances and their MMSE estimates (ROSIS dataset). (Top) HS image. (Bottom) MS image.}
\label{fig:Var_noise_pavia}
\end{figure}

This section considers another reference image (the high spatial
and high spectral image is a $128 \times 64 \times 93$ HS image
with very high spatial resolution of 1.3 m/pixel) acquired by the
Reflective Optics System Imaging Spectrometer (ROSIS) optical sensor
over the urban area of the University of Pavia, Italy. The flight was
operated by the Deutsches Zentrum f\"{u}r Luft- und Raumfahrt (DLR, the
German Aerospace Agency) in the framework of the
HySens project, managed and sponsored by the European
Union. This image was initially composed of $115$ bands
that have been reduced to $93$ bands after removing the water
vapor absorption bands (with spectral range from 0.43 to 0.86 $\mu$m).
This image has received a lot of attention in the remote sensing literature
\cite{Plaza2009,Tarabalka2010,Li2013}. The HS blurring kernel is the same
as in paragraph \ref{subsec:fuse_HS_MS} and the MS spectral response is a $4$-band
IKONOS-like reflectance spectral response. The noise level is defined by
SNR$_{1,\cdot}=35$dB for the first 43 bands and SNR$_{1,\cdot} = 30$dB for the
remaining 50 bands of the HS image. For MS image, SNR$_{2,\cdot}$ is 30dB for all bands.
The ground-truth, HS, MS and fusion results obtained with different algorithms
are displayed in Fig. \ref{fig:Pavia}. The corresponding image quality measures
are reported in Table \ref{tb:quality_Pavia}. The estimates of the noise variances are shown
in Fig. \ref{fig:Var_noise_pavia}. These results are in good agreement with the
performance obtained before.

\begin{table}
\centering
\caption{Performance of different fusion methods in terms of: RSNR (\lowercase{dB}), UIQI, SAM (deg), ERGAS and DD($\times 10^{-2}$) (ROSIS dataset).}
\label{tb:quality_Pavia}
\begin{tabular}{c|c|c|c|c|c|c}
\hline
Methods & RSNR  & UIQI & SAM  &ERGAS & DD& Time(s) \\
\hline
MAP \cite{Hardie2004}      & 26.58  & 0.9926 &  2.90& 1.36 & 3.61 &$\bs{1.5}$\\
Wavelet \cite{Zhang2009}  & 26.62  & 0.9925 & 2.87 & 1.35 & 3.60 &30\\
Proposed & $\bs{27.30}$  &$\bs{0.9933}$  & $\bs{2.60}$ & $\bs{1.24}$ & $\bs{3.27}$ &410\\
\hline
\end{tabular}
\end{table}

\subsection{Application to pansharpening}
\begin{figure}[htb]
\centering
    \subfigure{
    \label{fig:subfig:HS}
    \includegraphics[width=0.12\textwidth]{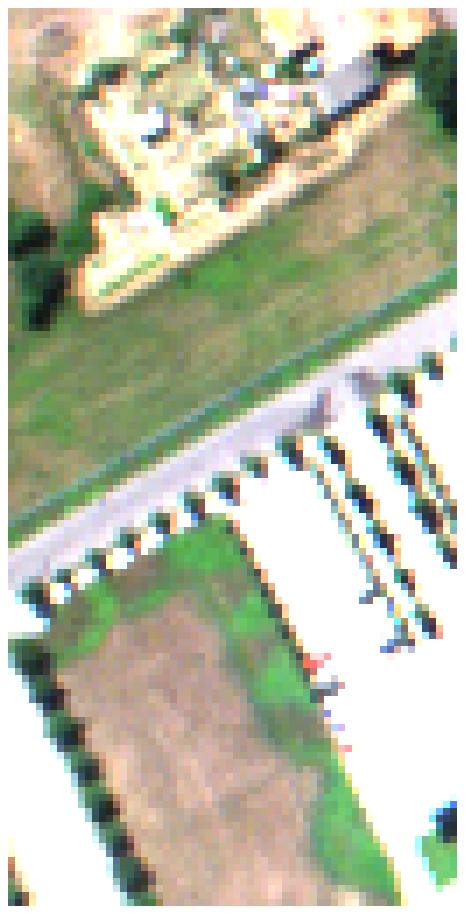}}
    \hspace{0in}
    \subfigure{
    \label{fig:subfig:MS}
    \includegraphics[width=0.12\textwidth]{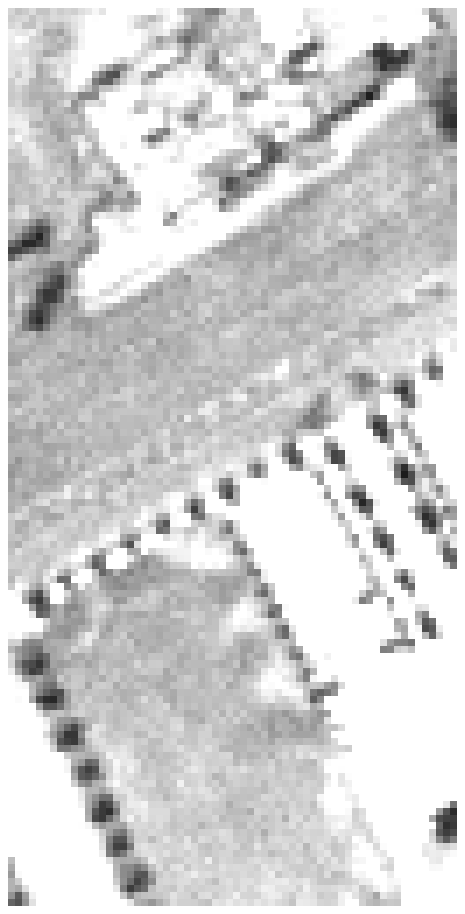}}
    \hspace{0in}
    \subfigure{
    \label{fig:subfig:Hardie}
    \includegraphics[width=0.12\textwidth]{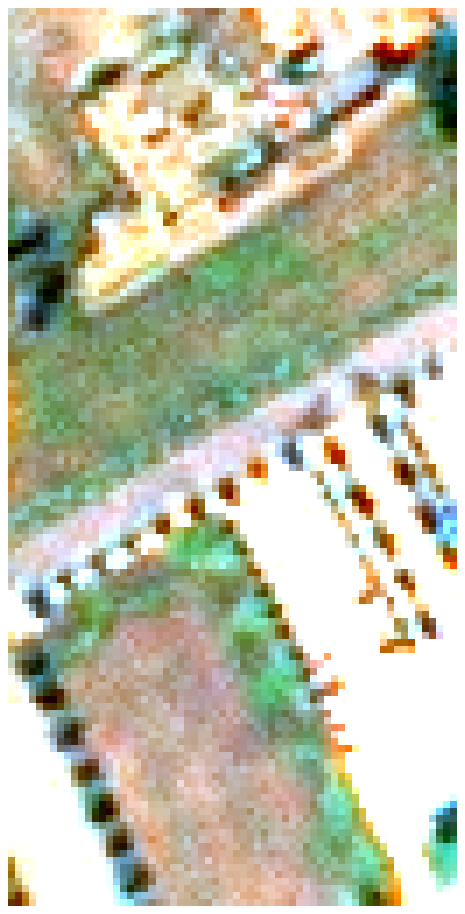}}\\
    \hspace{0in}
    \subfigure{
    \label{fig:subfig:Zhang}
    \includegraphics[width=0.12\textwidth]{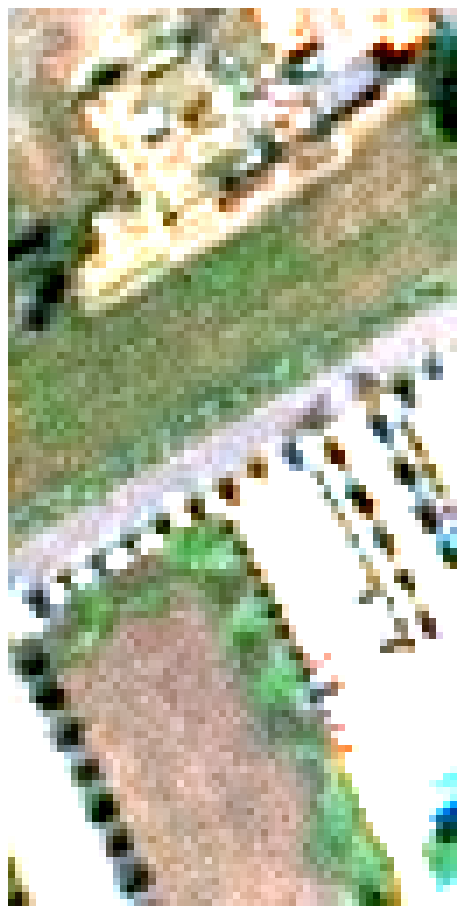}}
    \hspace{0in}
    \subfigure{
    \label{fig:subfig:HMC}
    \includegraphics[width=0.12\textwidth]{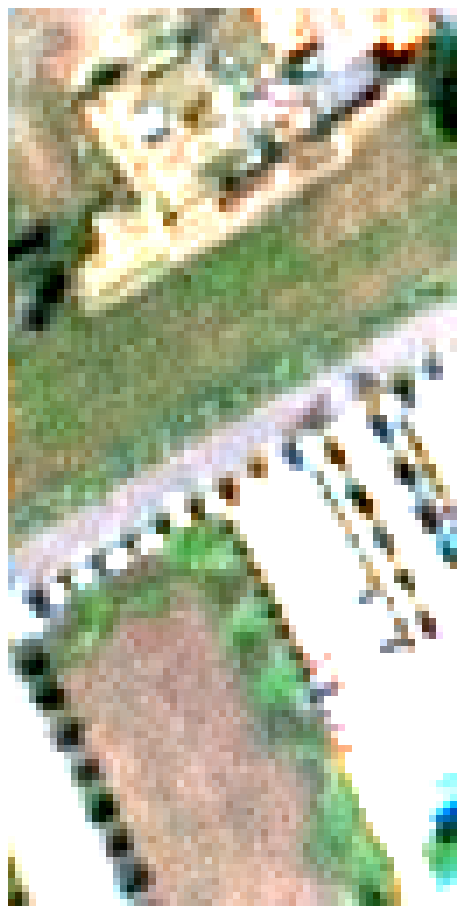}}
    \hspace{0in}
    \subfigure{
    \label{fig:subfig:Ref}
    \includegraphics[width=0.12\textwidth]{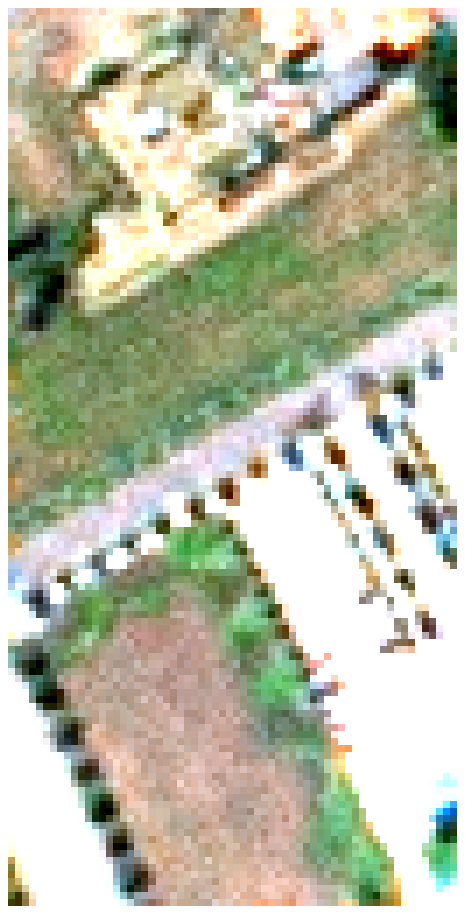}}
    \hspace{0in}
    \caption{ROSIS dataset: (Top left) Reference. (Top middle) MS Image. (Top right) Adaptive IHS \cite{Rahmani2010}.
     (Bottom left) MAP \cite{Hardie2004}. (Bottom middle) Wavelet MAP \cite{Zhang2009}. (Bottom right) Hamiltonian MCMC.}
    \label{fig:HS_PAN_Pavia}
\end{figure}

The proposed algorithm can also be used for pansharpening, which is
a quite important and popular application in the area of remote sensing.
In this section, we focus on fusing panchromatic and hyperspectral images (HS+PAN),
which is the extension of conventional pansharpening (MS+PAN). The HS image considered in this section
was used in paragraph \ref{subsec:Pavia} whereas the PAN image was obtained by averaging all the high resolution
HS bands. The SNR of the PAN image is $30$dB. Apart from \cite{Hardie2004,Zhang2009}, we also compare the results with
the method of \cite{Rahmani2010}, which proposes a popular pansharpening method. The results are displayed in Fig. \ref{fig:HS_PAN_Pavia} and
the quantitative results are reported in Table \ref{tb:quality_PAN_Pavia}. The proposed Bayesian
method still provides interesting results.

%Pansharpening is a particular case of multi-band fusion when the number of MS bands
%reduces to be 1.

\begin{table}
\centering
\caption{Performance of different fusion methods in terms of: RSNR (\lowercase{dB}), UIQI, SAM (deg), ERGAS and DD($\times 10^{-2}$) (ROSIS dataset).}
\label{tb:quality_PAN_Pavia}
\begin{tabular}{c|c|c|c|c|c|c}
\hline
Methods & RSNR  & UIQI & SAM  &ERGAS & DD& Time(s) \\
\hline
AIHS \cite{Rahmani2010}  & 16.69  & 0.9176 &  7.23& 4.24 & 9.99 &7.7\\
MAP \cite{Hardie2004}      & 17.54  & 0.9177&  6.55& 3.78& 8.78 &$\bs{1.4}$\\
Wavelet \cite{Zhang2009} & 18.03  & 0.9302 & 6.08 & 3.57 & 8.33 &26\\
Proposed & $\bs{18.23}$  &$\bs{0.9341}$  & $\bs{6.05}$ & $\bs{3.49}$ & $\bs{8.20}$ &387 \\
\hline
\end{tabular}
\end{table}

\section{Conclusions}
\label{sec:conclusions}
This paper proposed a hierarchical Bayesian model to fuse
multiple multi-band images with various spectral and spatial
resolutions. The image to be recovered was assumed to be degraded
according to physical transformations included within a forward
model. An appropriate prior distribution, that exploited geometrical
concepts encountered in spectral unmixing problems was proposed. The
resulting posterior distribution was efficiently sampled thanks to a
Hamiltonian Monte Carlo algorithm. Simulations conducted on
pseudo-real data showed that the proposed method competed with the
state-of-the-art techniques to fuse MS and HS images. These
experiments also illustrated the robustness of the proposed method
with respect to the misspecification of the forward model. Future work
includes the estimation of the parameters involved in the
forward model (e.g., the spatial and spectral responses of the
sensors) to obtain a fully unsupervised fusion
algorithm. The incorporation of spectral mixing constraints
for a possible improved spectral accuracy for the estimated high resolution
image would also deserve some attention.

\section*{Acknowledgments}
The authors would like to thank Dr. Paul Scheunders and Dr. Yifan
Zhang for sharing the codes of \cite{Zhang2009} and Jordi Inglada,
from Centre National d'\'Etudes Spatiales (CNES), for providing the
LANDSAT spectral responses used in the experiments. The authors also
acknowledge Prof. Jos\'e M. Bioucas Dias for valuable discussions
about this work that were handled during his visit in Toulouse
within the CIMI Labex.

\bibliographystyle{ieeetran}
%\bibliography{strings_all_ref,D:/qwei2/Dropbox/Latex/biblio_all}
\bibliography{strings_all_ref,biblio_all}

\begin{thebibliography}{10}
\providecommand{\url}[1]{#1}
\csname url@samestyle\endcsname
\providecommand{\newblock}{\relax}
\providecommand{\bibinfo}[2]{#2}
\providecommand{\BIBentrySTDinterwordspacing}{\spaceskip=0pt\relax}
\providecommand{\BIBentryALTinterwordstretchfactor}{4}
\providecommand{\BIBentryALTinterwordspacing}{\spaceskip=\fontdimen2\font plus
\BIBentryALTinterwordstretchfactor\fontdimen3\font minus
  \fontdimen4\font\relax}
\providecommand{\BIBforeignlanguage}[2]{{%
\expandafter\ifx\csname l@#1\endcsname\relax
\typeout{** WARNING: IEEEtran.bst: No hyphenation pattern has been}%
\typeout{** loaded for the language `#1'. Using the pattern for}%
\typeout{** the default language instead.}%
\else
\language=\csname l@#1\endcsname
\fi
#2}}
\providecommand{\BIBdecl}{\relax}
\BIBdecl

\bibitem{Amro2011survey}
I.~Amro, J.~Mateos, M.~Vega, R.~Molina, and A.~K. Katsaggelos, ``A survey of
  classical methods and new trends in pansharpening of multispectral images,''
  \emph{EURASIP J. Adv. Signal Process.}, vol. 2011, no.~1, pp. 1--22, 2011.

\bibitem{Dou2007general}
W.~Dou, Y.~Chen, X.~Li, and D.~Z. Sui, ``A general framework for component
  substitution image fusion: An implementation using the fast image fusion
  method,'' \emph{Comput. \& Geosci.}, vol.~33, no.~2, pp. 219--228, 2007.

\bibitem{Wald1999}
L.~Wald, ``Some terms of reference in data fusion,'' \emph{IEEE Trans. Geosci.
  and Remote Sens.}, vol.~37, no.~3, pp. 1190 --1193, May 1999.

\bibitem{Tu2004}
T.-M. Tu, P.~S. Huang, C.-L. Hung, and C.-P. Chang, ``A fast
  intensity-hue-saturation fusion technique with spectral adjustment for
  {IKONOS} imagery,'' \emph{IEEE Geosci. and Remote Sensing Lett.}, vol.~1,
  no.~4, pp. 309--312, 2004.

\bibitem{Aanaes2008}
H.~Aanaes, J.~Sveinsson, A.~Nielsen, T.~Bovith, and J.~Benediktsson,
  ``Model-based satellite image fusion,'' \emph{IEEE Trans. Geosci. and Remote
  Sens.}, vol.~46, no.~5, pp. 1336--1346, May 2008.

\bibitem{Joshi2010}
M.~Joshi and A.~Jalobeanu, ``{MAP} estimation for multiresolution fusion in
  remotely sensed images using an {IGMRF} prior model,'' \emph{IEEE Trans.
  Geosci. and Remote Sens.}, vol.~48, no.~3, pp. 1245--1255, March 2010.

\bibitem{Liu2012}
D.~Liu and P.~T. Boufounos, ``Dictionary learning based pan-sharpening,'' in
  \emph{Proc. IEEE Int. Conf. Acoust., Speech, and Signal Processing (ICASSP)},
  Kyoto, Japan, March 2012, pp. 2397--2400.

\bibitem{Manolakis2002}
D.~Manolakis and G.~Shaw, ``Detection algorithms for hyperspectral imaging
  applications,'' \emph{IEEE Signal Process. Mag.}, vol.~19, no.~1, pp. 29--43,
  jan 2002.

\bibitem{Chang2003}
{C.-I Chang}, \emph{Hyperspectral Imaging: Techniques for Spectral detection
  and classification}.\hskip 1em plus 0.5em minus 0.4em\relax New York: Kluwer,
  2003.

\bibitem{Bioucas2012}
J.~M. Bioucas-Dias, A.~Plaza, N.~Dobigeon, M.~Parente, Q.~Du, P.~Gader, and
  J.~Chanussot, ``Hyperspectral unmixing overview: Geometrical, statistical,
  and sparse regression-based approaches,'' \emph{IEEE J. Sel. Topics Appl.
  Earth Observations and Remote Sens.}, vol.~5, no.~2, pp. 354--379, 2012.

\bibitem{Cetin2009Merging}
M.~Cetin and N.~Musaoglu, ``Merging hyperspectral and panchromatic image data:
  qualitative and quantitative analysis,'' \emph{Int. J. Remote Sens.},
  vol.~30, no.~7, pp. 1779--1804, 2009.

\bibitem{Licciardi2012}
G.~A. Licciardi, M.~M. Khan, J.~Chanussot, A.~Montanvert, L.~Condat, and
  C.~Jutten, ``Fusion of hyperspectral and panchromatic images using
  multiresolution analysis and nonlinear pca band reduction,'' \emph{EURASIP J.
  Adv. Signal Process.}, vol. 2012, no.~1, pp. 1--17, 2012.

\bibitem{Moeller2009}
M.~Moeller, T.~Wittman, and A.~L. Bertozzi, ``A variational approach to
  hyperspectral image fusion,'' in \emph{Proc. SPIE Defense, Security, and
  Sensing}.\hskip 1em plus 0.5em minus 0.4em\relax International Society for
  Optics and Photonics, 2009, pp. 73\,341E--73\,341E.

\bibitem{Chisense2012}
C.~Chisense, J.~Engels, M.~Hahn, and E.~G{\"u}lch, ``Pansharpening of
  hyperspectral images in urban areas,'' in \emph{Proc. XXII Congr. of the Int.
  Society for Photogrammetry, Remote Sens.}, Melbourne, Australia, 2012.

\bibitem{Winter2002resolution}
M.~E. Winter and E.~Winter, ``Resolution enhancement of hyperspectral data,''
  in \emph{Proc. IEEE Aerospace Conference}, 2002, pp. 3--1523.

\bibitem{Chen2012super}
G.~Chen, S.-E. Qian, J.-P. Ardouin, and W.~Xie, ``Super-resolution of
  hyperspectral imagery using complex ridgelet transform,'' \emph{Int. J.
  Wavelets, Multiresolution Inf. Process.}, vol.~10, no.~03, 2012.

\bibitem{website:WV3}
D.~Inc., ``Worldview-3,''
  \url{http://www.satimagingcorp.com/satellite-sensors/WorldView3-DS-WV3-Web.pdf},
  Jan. 2013.

\bibitem{Ohgi2010}
N.~Ohgi, A.~Iwasaki, T.~Kawashima, and H.~Inada, ``Japanese hyper-multi
  spectral mission,'' in \emph{Proc. IEEE Int. Conf. Geosci. Remote Sens.
  (IGARSS)}, Honolulu, Hawaii, USA, July 2010, pp. 3756--3759.

\bibitem{Yokoya2013}
N.~Yokoya and A.~Iwasaki, ``Hyperspectral and multispectral data fusion mission
  on hyperspectral imager suite ({HISUI}),'' in \emph{Proc. IEEE Int. Conf.
  Geosci. Remote Sens. (IGARSS)}, Melbourne, Australia, July 2013, pp.
  4086--4089.

\bibitem{Shettigara1992}
V.~Shettigara, ``A generalized component substitution technique for spatial
  enhancement of multispectral images using a higher resolution data set,''
  \emph{Photogramm. Eng. Remote Sens.}, vol.~58, no.~5, pp. 561--567, 1992.

\bibitem{Zhou1998wavelet}
J.~Zhou, D.~Civco, and J.~Silander, ``A wavelet transform method to merge
  {L}andsat {TM} and {SPOT} panchromatic data,'' \emph{Int. J. Remote Sens.},
  vol.~19, no.~4, pp. 743--757, 1998.

\bibitem{Gonzalez2004fusion}
M.~Gonz{\'a}lez-Aud{\'\i}cana, J.~L. Saleta, R.~G. Catal{\'a}n, and
  R.~Garc{\'\i}a, ``Fusion of multispectral and panchromatic images using
  improved {IHS} and {PCA} mergers based on wavelet decomposition,'' \emph{IEEE
  Trans. Geosci. and Remote Sens.}, vol.~42, no.~6, pp. 1291--1299, 2004.

\bibitem{Hardie2004}
R.~C. Hardie, M.~T. Eismann, and G.~L. Wilson, ``{MAP} estimation for
  hyperspectral image resolution enhancement using an auxiliary sensor,''
  \emph{IEEE Trans. Image Process.}, vol.~13, no.~9, pp. 1174--1184, Sept.
  2004.

\bibitem{Zhang2009}
Y.~Zhang, S.~De~Backer, and P.~Scheunders, ``Noise-resistant wavelet-based
  {B}ayesian fusion of multispectral and hyperspectral images,'' \emph{IEEE
  Trans. Geosci. and Remote Sens.}, vol.~47, no.~11, pp. 3834 --3843, Nov.
  2009.

\bibitem{Zhang2012}
Y.~Zhang, A.~Duijster, and P.~Scheunders, ``A {B}ayesian restoration approach
  for hyperspectral images,'' \emph{IEEE Trans. Geosci. and Remote Sens.},
  vol.~50, no.~9, pp. 3453 --3462, Sep. 2012.

\bibitem{Eismann2004}
M.~T. Eismann and R.~C. Hardie, ``Application of the stochastic mixing model to
  hyperspectral resolution enhancement,'' \emph{IEEE Trans. Geosci. and Remote
  Sens.}, vol.~42, no.~9, pp. 1924--1933, Sept. 2004.

\bibitem{Eismann2005}
------, ``Hyperspectral resolution enhancement using high-resolution
  multispectral imagery with arbitrary response functions,'' \emph{IEEE Trans.
  Image Process.}, vol.~43, no.~3, pp. 455--465, March 2005.

\bibitem{Otazu2005}
X.~Otazu, M.~Gonzalez-Audicana, O.~Fors, and J.~Nunez, ``Introduction of sensor
  spectral response into image fusion methods. {A}pplication to wavelet-based
  methods,'' \emph{IEEE Trans. Geosci. and Remote Sens.}, vol.~43, no.~10, pp.
  2376--2385, 2005.

\bibitem{Joshi2006}
M.~V. Joshi, L.~Bruzzone, and S.~Chaudhuri, ``A model-based approach to
  multiresolution fusion in remotely sensed images,'' \emph{IEEE Trans. Geosci.
  and Remote Sens.}, vol.~44, no.~9, pp. 2549--2562, Sept. 2006.

\bibitem{Casella1992}
G.~Casella and E.~I. George, ``Explaining the {G}ibbs sampler,'' \emph{The
  American Statistician}, vol.~46, no.~3, pp. 167--174, 1992.

\bibitem{Hastings1970}
W.~K. Hastings, ``{M}onte {C}arlo sampling methods using {M}arkov chains and
  their applications,'' \emph{Biometrika}, vol.~57, no.~1, pp. 97--109, 1970.

\bibitem{Duane1987}
S.~Duane, A.~D. Kennedy, B.~J. Pendleton, and D.~Roweth, ``Hybrid {M}onte
  {C}arlo,'' \emph{Physics Lett. B}, vol. 195, no.~2, pp. 216--222, Sept. 1987.

\bibitem{Neal1993}
R.~M. Neal, ``Probabilistic inference using {M}arkov chain {M}onte {C}arlo
  methods,'' Dept. of Computer Science, University of Toronto, Tech. Rep.
  CRG-TR-93-1, Sept. 1993.

\bibitem{Neal2010}
------, ``{MCMC} using {H}amiltonian dynamics,'' \emph{Handbook of Markov Chain
  Monte Carlo}, vol.~54, pp. 113--162, 2010.

\bibitem{Fasbender2008}
D.~Fasbender, D.~Tuia, P.~Bogaert, and M.~Kanevski, ``Support-based
  implementation of {B}ayesian data fusion for spatial enhancement:
  Applications to {ASTER} thermal images,'' \emph{IEEE Geosci. and Remote
  Sensing Lett.}, vol.~5, no.~4, pp. 598--602, Oct. 2008.

\bibitem{Elbakary2008}
M.~Elbakary and M.~Alam, ``Superresolution construction of multispectral
  imagery based on local enhancement,'' \emph{IEEE Geosci. and Remote Sensing
  Lett.}, vol.~5, no.~2, pp. 276--279, April 2008.

\bibitem{Campbell2002}
J.~B. Campbell, \emph{Introduction to remote sensing}, 3rd~ed.\hskip 1em plus
  0.5em minus 0.4em\relax New-York, NY: Taylor \& Francis, 2002.

\bibitem{Jalobeanu2004}
A.~Jalobeanu, L.~Blanc-Feraud, and J.~Zerubia, ``An adaptive {G}aussian model
  for satellite image deblurring,'' \emph{IEEE Trans. Image Process.}, vol.~13,
  no.~4, pp. 613--621, 2004.

\bibitem{Duijster2009}
A.~Duijster, P.~Scheunders, and S.~De~Backer, ``Wavelet-based em algorithm for
  multispectral-image restoration,'' \emph{IEEE Trans. Geosci. and Remote
  Sens.}, vol.~47, no.~11, pp. 3892--3898, 2009.

\bibitem{Xu2011}
M.~Xu, H.~Chen, and P.~K. Varshney, ``An image fusion approach based on
  {M}arkov random fields,'' \emph{IEEE Trans. Geosci. and Remote Sens.},
  vol.~49, no.~12, pp. 5116--5127, 2011.

\bibitem{Dobigeon2008icassp}
N.~Dobigeon, J.-Y. Tourneret, and {A. O. Hero III}, ``Bayesian linear unmixing
  of hyperspectral images corrupted by colored {G}aussian noise with unknown
  covariance matrix,'' in \emph{Proc. IEEE Int. Conf. Acoust., Speech, and
  Signal Processing (ICASSP)}, Las Vegas, USA, March 2008, pp. 3433--3436.

\bibitem{Schultz1994}
R.~Schultz and R.~Stevenson, ``A {B}ayesian approach to image expansion for
  improved definition,'' \emph{IEEE Trans. Image Process.}, vol.~3, no.~3, pp.
  233--242, May 1994.

\bibitem{Chang1998}
C.-I. Chang, X.-L. Zhao, M.~L. Althouse, and J.~J. Pan, ``Least squares
  subspace projection approach to mixed pixel classification for hyperspectral
  images,'' \emph{IEEE Trans. Geosci. and Remote Sens.}, vol.~36, no.~3, pp.
  898--912, 1998.

\bibitem{Bioucas2008}
J.~M. Bioucas-Dias and J.~M. Nascimento, ``Hyperspectral subspace
  identification,'' \emph{IEEE Trans. Geosci. and Remote Sens.}, vol.~46,
  no.~8, pp. 2435--2445, 2008.

\bibitem{Hardie1997}
R.~C. Hardie, K.~J. Barnard, and E.~E. Armstrong, ``Joint {MAP} registration
  and high-resolution image estimation using a sequence of undersampled
  images,'' \emph{IEEE Trans. Image Process.}, vol.~6, no.~12, pp. 1621--1633,
  Dec. 1997.

\bibitem{Woods2006}
N.~A. Woods, N.~P. Galatsanos, and A.~K. Katsaggelos, ``Stochastic methods for
  joint registration, restoration, and interpolation of multiple undersampled
  images,'' \emph{IEEE Trans. Image Process.}, vol.~15, no.~1, pp. 201--213,
  Jan. 2006.

\bibitem{Qi2014_TechRep}
\BIBentryALTinterwordspacing
Q.~Wei, N.~Dobigeon, and J.-Y. Tourneret, ``Bayesian fusion of multi-band
  images,'' \emph{IRIT-ENSEEIHT, Tech. Report, Univ. of Toulouse}, 2014.
  [Online]. Available:
  \url{http://wei.perso.enseeiht.fr/papers/2014QiTechnicalReport.pdf}
\BIBentrySTDinterwordspacing

\bibitem{Punskaya2002}
E.~Punskaya, C.~Andrieu, A.~Doucet, and W.~Fitzgerald, ``{B}ayesian curve
  fitting using {M}{C}{M}{C} with applications to signal segmentation,''
  \emph{IEEE Trans. Signal Process.}, vol.~50, no.~3, pp. 747--758, March 2002.

\bibitem{Gelman2006prior}
A.~Gelman, ``Prior distributions for variance parameters in hierarchical models
  (comment on article by browne and draper),'' \emph{Bayesian analysis},
  vol.~1, no.~3, pp. 515--534, 2006.

\bibitem{Robert2007}
C.~P. Robert, \emph{The {B}ayesian Choice: from Decision-Theoretic Motivations
  to Computational Implementation}, 2nd~ed., ser. Springer Texts in
  Statistics.\hskip 1em plus 0.5em minus 0.4em\relax New York, NY, USA:
  Springer-Verlag, 2007.

\bibitem{Bidon2008}
S.~Bidon, O.~Besson, and J.-Y. Tourneret, ``The adaptive coherence estimator is
  the generalized likelihood ratio test for a class of heterogeneous
  environments,'' \emph{IEEE Signal Process. Lett.}, vol.~15, pp. 281--284,
  2008.

\bibitem{Bouriga2012}
M.~Bouriga and O.~F{\'e}ron, ``Estimation of covariance matrices based on
  hierarchical inverse-{W}ishart priors,'' \emph{J. of Stat. Planning and
  Inference}, 2012.

\bibitem{Zhang2012Generative}
H.~Zhang, Y.~Zhang, H.~Li, and T.~S. Huang, ``Generative {B}ayesian image super
  resolution with natural image prior,'' \emph{IEEE Trans. Image Process.},
  vol.~21, no.~9, pp. 4054--4067, 2012.

\bibitem{Orieux2012}
F.~Orieux, O.~F{\'e}ron, and J.-F. Giovannelli, ``Sampling high-dimensional
  {G}aussian distributions for general linear inverse problems,'' \emph{IEEE
  Signal Process. Lett.}, vol.~19, no.~5, pp. 251--254, 2012.

\bibitem{David2012}
D.~Ceperley, Y.~Chen, R.~V. Craiu, X.-L. Meng, A.~Mira, and J.~Rosenthal,
  ``Challenges and advances in high dimensional and high complexity monte carlo
  computation and theory,'' in \emph{Banff International Research Station},
  2012.

\bibitem{Green1998imaging}
R.~O. Green, M.~L. Eastwood, C.~M. Sarture, T.~G. Chrien, M.~Aronsson, B.~J.
  Chippendale, J.~A. Faust, B.~E. Pavri, C.~J. Chovit, M.~Solis \emph{et~al.},
  ``Imaging spectroscopy and the airborne visible/infrared imaging spectrometer
  ({AVIRIS}),'' \emph{Remote Sens. of Environment}, vol.~65, no.~3, pp.
  227--248, 1998.

\bibitem{Roberts2007}
\BIBentryALTinterwordspacing
G.~O. Roberts and J.~S. Rosenthal, ``\BIBforeignlanguage{English}{Coupling and
  ergodicity of adaptive {M}arkov {C}hain {M}onte {C}arlo algorithms},''
  \emph{\BIBforeignlanguage{English}{J. of Appl. Probability}}, vol.~44, no.~2,
  pp. pp. 458--475, 2007. [Online]. Available:
  \url{http://www.jstor.org/stable/27595854}
\BIBentrySTDinterwordspacing

\bibitem{Wang2002}
Z.~Wang and A.~C. Bovik, ``A universal image quality index,'' \emph{IEEE Signal
  Process. Lett.}, vol.~9, no.~3, pp. 81--84, 2002.

\bibitem{Wald2000}
L.~Wald, ``Quality of high resolution synthesised images: Is there a simple
  criterion?'' in \emph{Proc. Int. Conf. Fusion of Earth Data}, Nice, France,
  Jan 2000, pp. 99--103.

\bibitem{Plaza2009}
A.~Plaza, J.~A. Benediktsson, J.~W. Boardman, J.~Brazile, L.~Bruzzone,
  G.~Camps-Valls, J.~Chanussot, M.~Fauvel, P.~Gamba, A.~Gualtieri,
  M.~Marconcini, J.~C. Tilton, and G.~Trianni, ``Recent advances in techniques
  for hyperspectral image processing,'' \emph{Remote Sensing of Environment},
  vol. 113, Supplement 1, pp. S110--S122, 2009.

\bibitem{Tarabalka2010}
Y.~Tarabalka, M.~Fauvel, J.~Chanussot, and J.~Benediktsson, ``{SVM}- and
  {MRF}-based method for accurate classification of hyperspectral images,''
  \emph{IEEE Trans. Geosci. and Remote Sens.}, vol.~7, no.~4, pp. 736--740,
  2010.

\bibitem{Li2013}
J.~Li, J.~M. Bioucas-Dias, and A.~Plaza, ``Spectral-spatial classification of
  hyperspectral data using loopy belief propagation and active learning,''
  \emph{IEEE Trans. Geosci. and Remote Sens.}, vol.~51, no.~2, pp. 844--856,
  2013.

\bibitem{Rahmani2010}
S.~Rahmani, M.~Strait, D.~Merkurjev, M.~Moeller, and T.~Wittman, ``An adaptive
  {IHS} pan-sharpening method,'' \emph{IEEE Geosci. and Remote Sensing Lett.},
  vol.~7, no.~4, pp. 746--750, 2010.

\end{thebibliography}
%\bibliography{D:/qwei2/Bayesian fusion/strings_all_ref,D:/qwei2/Bayesian fusion/biblio}
\end{document}